\RequirePackage{fix-cm}
\documentclass[smallextended]{svjour3}

\usepackage{natbib}
\usepackage{graphicx}
\usepackage{amssymb}
\usepackage{mathtools}

\usepackage{breqn}
\usepackage{bm}
\usepackage{color}
\usepackage{tikz}
\usepackage{subfigure}
\usepackage{url}

\smartqed  

\def\dom{{\rm dom}}

\let\originalleft\left
\let\originalright\right
\renewcommand{\left}{\mathopen{}\mathclose\bgroup\originalleft}
\renewcommand{\right}{\aftergroup\egroup\originalright}

\begin{document}
\date{Accepted for publication in Data Mining and Knowledge Discovery on December 10, 2015}
\journalname{Data Mining and Knowledge Discovery}
\title{Characterizing Concept Drift}
\author{Geoffrey I.\ Webb \and Roy Hyde \and Hong Cao \and Hai Long Nguyen \and Francois Petitjean}
\institute{G.~I.~Webb \at
       Faculty of Information Technology\\
       Monash University\\
       Clayton, Vic 3800, Australia\\
       \email{geoff.webb@monash.edu}
       \and R.~Hyde \at
       Faculty of Information Technology\\
       Monash University\\
       Clayton, Vic 3800, Australia\\
       \email{roy.hyde@alumni.monash.edu}
       \and H.~Cao \at
       McLaren Applied Technologies Pte Ltd APAC\\
       Suntec Tower One, Singapore 038987\\
       \email{hong.cao@mclaren.com}
       \and H.~L.~Nguyen \at
       McLaren Applied Technologies Pte Ltd APAC\\
       Suntec Tower One, Singapore 038987\\
       \email{long.nguyen@mclaren.com}
       \and F.~Petitjean \at
       Faculty of Information Technology\\
       Monash University\\
       Clayton, Vic 3800, Australia\\
       \email{francois.petitjean@monash.edu}
       }

\maketitle

\begin{abstract}
Most machine learning models are static, but the world is dynamic, and increasing online deployment of learned models gives increasing urgency to the development of efficient and effective mechanisms to address learning in the context of non-stationary distributions, or as it is commonly called {\em concept drift}.  However, the key issue of characterizing the  different types of drift that can occur has not previously been subjected to rigorous definition and analysis.  In particular, while some qualitative drift categorizations have been proposed, few have been formally defined, and the quantitative descriptions required for precise and objective understanding of learner performance have not existed.  We present the first comprehensive framework for quantitative analysis of drift.  This supports the development of the first comprehensive set of formal definitions of types of concept drift. The formal definitions clarify ambiguities and identify gaps in previous definitions, giving rise to a new comprehensive taxonomy of concept drift types and a solid foundation for research into mechanisms to detect and address concept drift.

\end{abstract}



\section{Introduction}

Most machine learning systems operate in batch mode. They analyze a set of historical data and then develop a model that reflects the world as it was when the model was formed.  But the world is dynamic, and the complex distributions that a model models are likely to be non-stationary and thus to change over time, leading to deteriorating model performance.  To address this problem it is necessary to develop mechanisms for detecting and handling concept drift.
To this end there has been much work on identifying types of concept drift \citep{presenceofconceptdrift,Gama09,LearningUnderConceptDrift,StreamingData+ConceptDrift+Imbalance,UnifyingView,SurveyConceptDriftAdaptation}.  While many types have been identified, most definitions have been qualitative, many of the definitions have been informal and despite efforts to introduce standardization \citep{UnifyingView}, the terminology has been non-uniform and inconsistently applied.

We argue that quantitative measures of concept drift are essential for the development of a detailed understanding of the problems confronted by attempts to detect and address non-stationary distributions, and to assess the relative capacity of techniques that may be proposed. While qualitative measures are necessarily subjective and imprecise, quantitative measures can be both objective and exact.

We propose some core quantitative measures and provide a framework for developing further such measures.  This framework also provides the foundation for formal definitions of qualitative categorizations of types of drift.
We exploit this to develop precise formal definitions for the key types of concept drift previously mentioned in the literature and provide a comprehensive taxonomy thereof.  

Our formal definitions allow some of the controversies in the field to be resolved, such as whether the distinction between incremental and gradual drift is meaningful and if so exactly how the two categories of drift relate to one another.  
\begin{figure}
\tabcolsep=.75\tabcolsep
\begin{tabular}{ccc}
\hspace*{-10pt}\includegraphics[width=.32\textwidth]{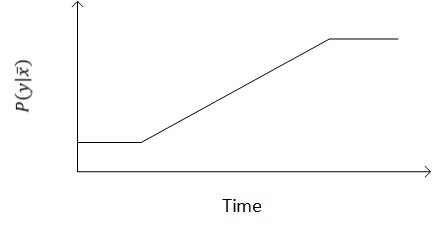}&
\hspace*{-10pt}\includegraphics[width=.32\textwidth]{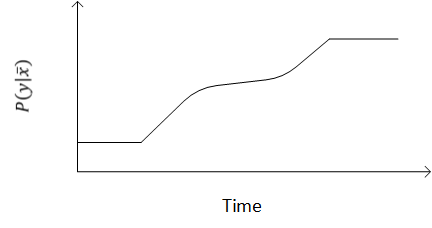}&
\hspace*{-10pt}\includegraphics[width=.32\textwidth]{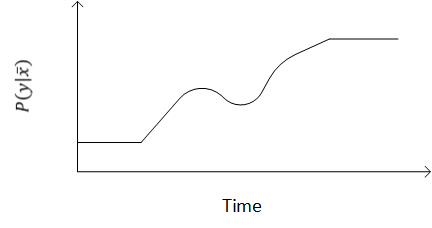}
\\[-3pt]
    
\hspace*{-10pt}\includegraphics[width=.32\textwidth]{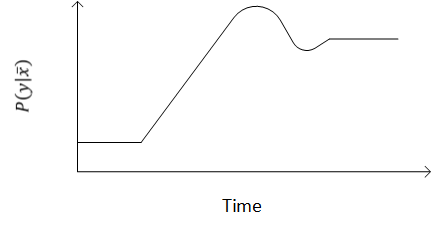}&
\hspace*{-10pt}\includegraphics[width=.32\textwidth]{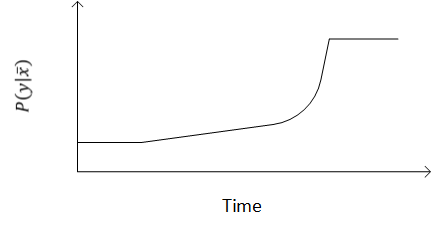}&
\hspace*{-10pt}\includegraphics[width=.32\textwidth]{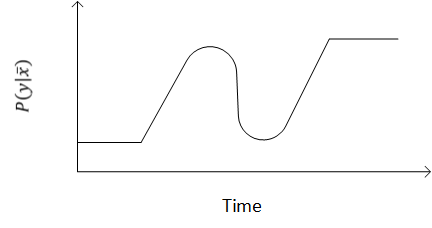}\\[-3pt]
  
   \end{tabular}
\caption{Examples of types of concept drift with respect to which previous definitions are ambiguous as to which are gradual.}
    \label{fig:gradinc}
\end{figure}%
They also remove the ambiguity inherent in many current definitions, as illustrated by the inability of previous definitions of {\em gradual drift\/} to clearly distinguish which of the forms of drift illustrated in Fig.~\ref{fig:gradinc} are gradual and which are not. Most importantly, they provide a solid foundation for the research community to develop new technologies to detect, characterize and resolve concept drift, and design new machine learning techniques that are robust to a greater diversity of drift types.

Much of the work on concept drift has concentrated on classification \citep{AWE,Zhang2008,Masud2011}.  However, it is also an issue in other forms of learned model that are deployed for extended periods of time, and of especial relevance to all forms of stream mining including clustering \citep{aggarwal2003framework}, association discovery \citep{jiang2006research} and outlier detection \citep{subramaniam2006online}.  While the issues relating to drift subject, raised in Section~\ref{subsec:DSubject}, are specific to classification models, the rest of the paper is broadly applicable across many forms of model.

This paper does not directly address issues of how to detect and handle concept drift from data, for which interested readers are encouraged to read the numerous existing excellent tutorials and surveys \citep{Bifet2011PAKDD,Nguyen2014,Gama09,Aggarwal09,Gaber05, Krempl2014survey} for the current state of the art and the open challenges.  We leave it to future research to apply the new conceptual tools that we have developed to these important problems.

\section{Initial Definitions}

Before we can discuss our new quantitative measures and formal definitions for different types of drift, we must define the terms and concepts we use to create them.  We couch our analysis in the context of data streams, but note that the fundamental issues and techniques are applicable to any context in which a model may be learned from historical data and applied in future contexts.

We first discuss data streams and their analysis.  We then present our definition of a concept and the common functions we have used to demonstrate the characterization of differences between concepts.

\subsection{Data Streams}
\label{subsec:DataStream}
We define a data stream as a data set in which the data elements have time stamps.  In a standard application, a system only has access to data with time stamps prior to a specific point of time, but the models to be developed must be applied to data elements with subsequent time stamps.

The process that generates the stream can be considered to be a random variable $\mathcal X$ from which the objects $o\in \dom(\mathcal X)$ are drawn at random, where $\dom(\cdot)$ denotes the domain of a random variable.  For classification learning, we need to distinguish a class variable or {\em label}, $y\in \dom(Y)$, where $Y$ denotes a random variable over class labels, and the covariates $x\in \dom(X)$, where $X$ denotes a random variable over vectors of attribute values.  In this case, ${\mathcal X}$ represents the joint distribution $XY$ and $o$ represents a pair $\left< x,y\right>$.
We provide a summary of the key symbols used in this paper in Table~\ref{tab:symbols}.

\begin{table}[t]
\begin{center}
\begin{tabular}{|l|p{2.65in}|l|}
\hline
\textbf{Symbols}& 
\textbf{Represents}& 
\textbf{Scope} \\
\hline
{$a, b$}& 
Concept index& 
$\mathbb{N}_{{>}0}$ \\
\hline
{$t, u, w$}& 
Points of time& 
$\mathbb{R}_{{>}0}$ \\
\hline
{$m$}& 
Durations& 
$\mathbb{R}_{{\geq}0}$ \\
\hline
{$i, j, k, n$}& 
General purpose non-negative integers& 
$\mathbb{N}$ \\
\hline
{$\phi$}& 
Minimum duration for concept stability
& 
$\mathbb{R}_{{\geq}0}$  \\
\hline
{$\delta$}&
Maximum duration for abrupt drift
& 
$\mathbb{R}_{{\geq}0}$  \\
\hline
{$\beta$}&
Maximum duration for drift to and from a concept to be considered blip drift
& 
$\mathbb{R}_{{\geq}0}$  \\
\hline
{$\gamma$}&
Minor/Major drift threshold
& 
$\mathbb{R}_{{>}0}$  \\
\hline
$\nu$& A drift period used in defining gradual drift.& 
$\mathbb{R}_{{>}0}$ 
\\\hline
$\mu$ & A maximum allowed difference between concepts over time period $\nu$ during a period of drift for the drift to be considered gradual.& 
$\mathbb{R}_{{>}0}$ 
\\\hline
$\mathcal X$&
A random variable over objects& 
Stream dependent \\
\hline
$o$&
An object& 
$\dom(\mathcal X)$ \\
\hline
$X$&
A random variable over covariates& 
Stream dependent \\
\hline
{$x$}&
A covariate& 
$\dom(X)$ \\
\hline
$Y$&
A random variable over class labels& 
Stream dependent \\
\hline
$y$&
Class label& 
$\dom(Y)$ \\
\hline
\end{tabular}


\caption{List of symbols used.}
\label{tab:symbols}
\end{center}
\end{table}


A learning algorithm analyses the training data to create a model for future test data. 

In the context of classification learning, we use $P(Y)$ to denote the prior probability distribution over the class labels, $P(X)$ to denote the prior probability distribution over covariates, $P(X,Y)$  to denote the joint probability distribution over objects and class labels, $P(Y\mid X)$ to denote the likelihood distribution over class labels given objects and $P(X\mid Y)$ to denote the posterior probability distribution over covariates given class labels.  

In order to reference the probability distribution at a particular time we add a time subscript, such as $P_t(\mathcal X)$, to denote a probability distribution at time $t$. 

We allow for the possibility of continuous time, as it is more general.  However, our definitions do not require modification if time is restricted to discrete units.

\subsection{Concepts and concept drift}

A formal definition of {\em concept} is a prerequisite for formal characterizations of {\em concept drift}.

The classical definition of a concept is by extension---a concept is a set of objects \citep{michalski1983theory,angluin1988queries}.  Two variants of a definition by extension are possible, by \emph{type} or \emph{token} \citep{typetoken}.  A definition by type defines a concept as a set of vectors of $X$ values such that any object with any vector of values in the set belongs to the concept, or equivalently, as a function $X\rightarrow Y$.  Such a definition does not allow that different objects with identical attribute values might belong to different concepts.  However, many applications of machine learning require that there be a many-to-many mapping from $X$ values to $Y$ values rather than the many-to-one mapping that such a definition implies.  The alternative definition by extension defines a concept as a set of object instances.  This is coherent, but does not seem useful in the context of stream learning, because in many applications each object instance will appear only once.  Even in a situation where a patient is diagnosed on multiple occasions, in some sense the object being classified is the patient presentation, rather than the patient, as they may have a condition on one presentation and not another.  As the sets of objects at different times will not overlap in such a case it is difficult to see how one might assess whether a concept has changed under such a definition.

It is perhaps for this reason that recent concept drift literature has instead given a probabilistic definition of {\em concept\/}   \citep{kuncheva2004classifier}. 
\cite{StreamingData+ConceptDrift+Imbalance} define a concept as the prior class probabilities $P(Y)$ and class conditional probabilities $P(X\mid Y)$.  As $P(Y)$ and $P(X\mid Y)$ uniquely determines the joint distribution $P(X,Y)$ and vice versa, this is equivalent to defining a concept as the joint distribution $P(X,Y)$, as proposed by \cite{SurveyConceptDriftAdaptation}.  For this reason we adopt Gama et.\ al.'s (\citeyear{SurveyConceptDriftAdaptation}) definition for stream classification.

\begin{equation}
Concept = P\left(X, Y\right).
\end{equation}
Beyond classification learning (i.e. in the unsupervised case, where there is no special class attribute) this is simply:
\begin{equation} \label{eq:concept}
Concept = P\left(\mathcal X\right).
\end{equation}
In the context of a data stream, we need to recognize that concepts may change over time. To this end we  define the concept at a particular time $t$ as  
\begin{equation} \label{eq:ConceptTime}
P_{t}\left(\mathcal X\right).
\end{equation}

Concept drift occurs between times $t$ and $u$ when the distributions change, 
\begin{equation}\label{eq:drift-definition}
P_t({\mathcal X})\neq P_u({\mathcal X}).
\end{equation}
Concept drift is the stream learning counterpart of {\em dataset shift\/} from batch learning \citep{quionero2009dataset,UnifyingView}.

\section{Quantitative measures of drift}

There has been little prior work on quantifying concept drift.  
\cite{bartlett2000learning} define a measure they call \emph{drift rate}.  However, they use a limited definition of a concept as a function  $f: X\rightarrow Y$. They define the drift rate at time $t$ as $P(f_t\neq f_{t+1})$. \cite{minku2009using} provide a similar measure that they call \emph{severity}, the primary difference being that they model drift as occurring between periods of concept stability and define the severity as the proportion of the instance space for which the class labels change between successive stable concepts.
\cite{kosina2010drift} investigate the use of these measures for detecting drift.
These definitions are limited in assuming that there is a many to one mapping from $X$ to $Y$ values. The earlier of these definitions is further limited in assuming that time is discrete, and appears to assume that $P_t(X)=P_{t+1}(X)$, as it is otherwise unclear over which of these distributions the probability of the inequality is defined.

We here provide quantitative measures of concept drift that allow that concepts may be probabilistic relationships between $X$ and $Y$ values.

Quantifying the degree of difference between two points of time is a key characterization of any concept drift.  We call this \textit{drift magnitude}.  However, the appropriate function for measuring drift magnitude may change from domain to domain.  Rather than specifying which measure of distance between distributions should be used, our definitions refer to an unspecified distribution distance function:
\begin{equation} \label{eq:DifFunction}
D( t,t{+}m ).
\end{equation}
This function returns a non-negative value indicating the magnitude of the difference in the concepts at times $t$ and $t{+}m$.  Note that while we only use the times as parameters to this function, it is the concepts at those times between which the distance is judged.

At this dawn of the exploration and analysis of quantitative characterization of concept drift, it is not clear what properties are desirable of a measure of distance between concepts.  Examples of distance functions that might be used include Kullback-Leibler Divergence \citep{kullback1951information} and Hellinger Distance \citep{HellingerDistance}.

It may be desirable to use a metric as
\begin{itemize}
\item it is hard to understand what sense might be made of negative distances;
\item it seems credible that the distance between concept A and B should be the same as the distance between B and A; and
\item the triangle inequality seems desirable if one is to consider the evolution of concept drift over a number of concepts.  It seems implausible that the distance from concept A to concept B should exceed the sum of the distances from A to an intermediate concept C and from C to B.
\end{itemize}
For this reason, in this paper we use
Hellinger Distance in our case study.

In the context of classification, it will also be relevant to use all of measures of distance between the joint distributions $P_t(X,Y)$ and $P_u(X, Y)$; the $X$ distributions, $P_t(X)$ and $P_u(X)$; the class distributions $P_t(Y)$ and $P_u(Y)$; the posterior distributions, $P_t(Y\mid X)$ and $P_u(Y\mid X)$; and the likelihood distributions, $P_t(X\mid Y)$ and $P_u(X\mid Y)$.

Given a measure of drift distance it is straightforward to define a set of useful quantitative measures of concept drift.

The first of these is  the \textit{magnitude} of a drift, which is simply the distance between the concepts at the start and end of the period of drift.  The magnitude of drift between times $t$ and $u$ is
\begin{equation}
Magnitude_{t,u}=D(t, u).
\end{equation}
The magnitude will greatly impact the manner in which a learner should respond to the drift.  

Another critical measure of drift is  \textit{drift duration}, the elapsed time over which a period of drift occurs.  The duration of a drift starting at time $t$ and ending at time $u$ is
\begin{equation}\label{eq:DriftDuration}
Duration_{t,u}=u-t.
\end{equation}

Another aspect is the length of the path that the drift traverses during a period of drift. Path length is defined as the cumulative deviation observed during the period of drift. Note that this quantity is bounded below by the drift magnitude over the period (assuming $D$ is a metric), with the latter being the path length observed if the drift observed was monotone (see the upper left plot of Figure 1). The path length of a drift between times $t$ and $u$ is
\begin{equation}
PathLen_{t,u}=\lim_{n\to\infty}\sum_{k=0}^{n-1}D\left(t+\frac{k}{n}(u-t),t+\frac{k{+}1}{n}(u-t)\right).
\end{equation}
The path length provides another means of quantifying differences between drifts.  For example, the gradual and incremental drifts depicted in the upper left and upper center plots of Figure \ref{fig:gradinc} both have the same magnitude, but the latter has greater path length.

A further important quantitative measure is \textit{drift rate}, which quantifies how fast the distribution is changing at time $t$.
\begin{equation}
Rate_t=\lim_{n\to\infty}n D\left(t{-}0.5{/}n,t{+}0.5{/}n\right).
\end{equation}

The average drift rate between times $t$ and $u$ is thus 
\begin{equation}
PathLen_{t,u}/\left(u-t\right).
\end{equation}

In the context of classification, for different purposes it may be useful to apply these measures with respect to any of the joint ($P(XY)$), $P(X)$, $P(Y)$, posterior ($P(Y\mid X)$) or likelihood ($P(X\mid Y)$) distributions.

\section{Qualitative characterization of drift types}


There is a substantial literature discussing types of concept drift, with terms such as \emph{abrupt drift} given informal definitions.

Our formal analysis of these definitions leads to the conclusion that they often make an implicit assumption that drift occurs over discrete periods of time that are bounded before and after by periods without drift.  The concept of \emph{abrupt drift} \citep{ProblemOfConceptDrift,StreamingData+ConceptDrift+Imbalance,LearningUnderConceptDrift,ConceptDriftProcessMining,TrackingDriftTypes,OnlineEnsembleLearningConceptDrift,ConceptDriftScenarioReview,brzezinski2010mining,AUE} illustrates this implicit assumption. Abrupt drift seems to have been intended to mean a change in distributions that occurs instantaneously or near instantaneously.  Although we are the first to introduce the notion of drift magnitude, we assume that  the notion of abrupt drift of small magnitude is coherent.  If this is the case, and it is allowed that abrupt drift may be immediately preceded and followed by periods of drift, then it is difficult to see how it can be possible to sensibly distinguish between periods of repeated abrupt drift and other extended periods of drift.

\subsection{Functions describing points in time in the stream}
\label{subsec:functions}

To aid the creation of formal definitions of the standard  qualitative characterizations of concept drift, we provide a formal model of this notion that a stream can be considered as being composed of a number of periods of time during which there are stable concepts, interspersed by periods of instability and concept change.  We define a period of concept stability as any interval $[t, u]$ such that $u\geq t{+}\phi$ and $\forall_{m\in (0,u{-}t]}\, D\left(t,t{+}m\right)=0$, where $\phi$ is the minimum time period over which a concept must remain invariant in order to be assessed stable.  We leave $\phi$ unspecified, as appropriate values may differ from domain to domain.

The function $S_a$ returns the starting time and the function  $E_a$ returns the ending time of the $a^{th}$ stable concept in a stream.
\begin{equation}S_a=\left\{\begin{array}{ll}\min\{t\mid \forall_{m{\in}(0,\phi]}\, D(t,t{+}m)=0\}&\vert\, a=1\\
\min\{t\mid t>E_{a-1} \wedge \forall_{m{\in}(0,\phi]}\, D(t,t{+}m)=0\}&\vert\, a>1
\end{array}\right.
\end{equation}
\begin{equation}E_a=\max\{t\mid \forall_{m\in(0,t{-}S_a]}\, D(S_a,S_a{+}m)=0\}.
\end{equation}

\begin{figure}
\begin{center}
\includegraphics[width=\linewidth,clip=true,trim=4pt 2pt 27pt 0pt]{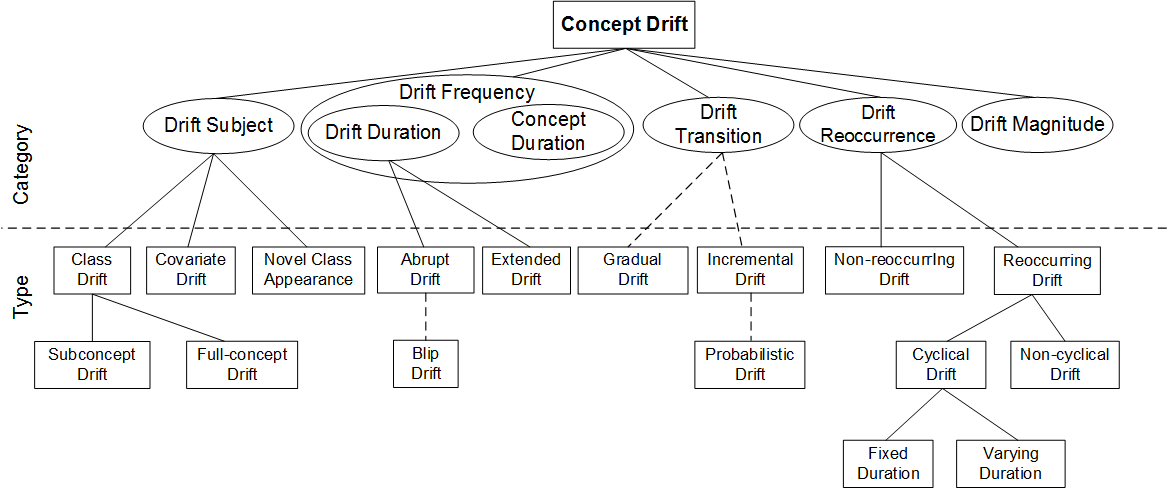}
\end{center}
\caption{A Taxonomy of Drift Categories}
\em Solid lines indicate that the child concepts form a complete set of mutually exclusive alternatives. For example, Drift Duration must be either Extended or Abrupt. Dashed~lines mean that the child concepts are not mutually exclusive.
\label{fig:diagrams}
\end{figure}

In this section we provide a taxonomy of categories of concept drift.  For each category defined we provide a formal definition. The taxonomy is summarized in Fig. \ref{fig:diagrams}.

\subsection{Drift Subject}
\label{subsec:DSubject}


In the context of classification learning, \citet{ChangingPopulations} observe that any of the following aspects of a joint distribution might change over time, $P(Y)$, $P(Y\mid X)$ or $P(X\mid Y)$. To this one can also add $P(X)$ \citep{ProblemOfConceptDrift}.  It should be noted that such changes are inter-related.  For example, $P(Y)$ cannot change without either $P(Y\mid X)$ or $P(X)$ changing.  It is also worth noting that a change over time in any of these aspects of a distribution requires a change in $P(X,Y)$, and hence all are captured by our definition of concept drift (\ref{eq:drift-definition}).

\citet{ChangingPopulations} suggest that in the context of classification learning $P(Y\mid X)$ is the most important of these drift subjects, as such a change will necessitate an update of a model if it is to maintain accuracy.  \cite{ProblemOfConceptDrift} argue that changes in $P(X)$ are also important, as an increase in the frequency of specific types of cases may increase the importance of accuracy in classifying those cases.

\subsubsection{Class drift}
\label{subsubsec:ClassDrift}
{\em Class drift}, often called {\em real concept drift\/} or \emph{prior probability shift}, occurs when the posterior class probabilities  $P(Y\mid X)$ change over time \citep{ProblemOfConceptDrift,StreamingData+ConceptDrift+Imbalance,UnifyingView,SurveyConceptDriftAdaptation}. For example, tax law may change over time and hence the set of tax-payer attributes that are associated with the class {\em compliant\/} will change.

Formally, class drift occurs between times $t$ and $u$ whenever
\begin{equation}
P_t(Y\mid X)\neq P_u(Y\mid X).\label{eq:ClassDrift}
\end{equation}

\cite{UnifyingView} define \emph{prior probability shift} which adds two further constraints to Eq.~\ref{eq:ClassDrift}:
\begin{enumerate}
\item that it only occurs when the covariates ($X$ variables) are causally determined by the class, and
\item that it only occurs when $P_t(X\mid Y)= P_u(X\mid Y)$.
\end{enumerate}
However, the value of the first constraint is questionable, as it is likely that in many real world learning tasks some $X$ variables have causal influence on the class, others are causally influenced by it, and still others simply share common causes. Further, for many learning systems, the direction of the causal relationships are not relevant, all that is utilized is the correlation between the covariates and the class.

The second constraint also seems needlessly strong.  To illustrate this take the following simple example.  Suppose:
\begin{itemize}
\item the covariates are types of product, with values $a$ or $b$;
\item the probability of each product remains $0.5$ throughout; 
\item the class is whether the product has a defect ($Y{=}d$) or not;
\item initially, at time $t$, $P_t(Y{=}d\mid X{=}a)=P_t(Y{=}d\mid X{=}b)=0.1$;
\item subsequent, at time $u$, product $a$ is improved and the probability of a defect is halved, $P_u(Y{=}d\mid X{=}a)=0.05$, while $P_u(Y{=}d\mid X{=}b)=0.1$ remains unchanged.
\end{itemize}
\sloppy In this case, because $P_u(X{=}a, Y{=}d)=0.025$ while $P_t(X{=}a, Y{=}d)=P_u(X{=}b, Y{=}d)=P_t(X{=}a, Y{=}d)=0.05$ we get $P_t(X{=}a\mid Y{=}d)=0.5\neq P_u(X{=}a\mid Y{=}d)=0.\dot{3}$.
This seems like a straightforward example of class drift, but is excluded by the above second constraint.

As an alternative to the second constraint, we propose the term \emph{pure class drift} to capture situations where it is only the posterior probability that is changing, defined as:
\begin{equation}
P_t(Y\mid X)\neq P_u(Y\mid X) \wedge P_t(X)= P_u(X).
\end{equation}

Class drift can be further divided into two sub-types that depend on the scope of the drift. {\em Drift scope}, also referred to as {\em drift severity\/} \citep{OnlineEnsembleLearningConceptDrift}, refers to the proportion of the domain of $X$ for which $P(Y\mid X)$ changes, 
\begin{equation}\label{eq:PureClassDrift}
\{x\in {\rm Dom}(X)\mid \exists_y\, P_t(Y{=}y\mid X{=}x)\neq P_u(Y{=}y\mid X{=}x)\}.
\end{equation}
Drift scope impacts the ease of detecting changes in the stream and affects how much of a model needs to be updated. 

{\em Subconcept drift}, also referred to as {\em intersected drift\/} \citep{OnlineEnsembleLearningConceptDrift}, is where the drift scope is limited to a subspace of $\dom(X)$. For example, a data stream that deals with financial records may have a class called `fraud', among others. If a new form of fraud is developed, the conditional probabilities of the fraud class occurring will change, but only for those contexts that relate to the new form of fraud. Whilst this is happening, cases that involve other forms of fraud could remain the same, and may continue in the same way as before.

Subconcept drift can be defined as:
\begin{multline} \label{eq:subconcept}
P_t(Y\mid X)\neq P_u(Y\mid X)\\
\wedge\,\exists_{x{\in}{\rm Dom}(X)}\,\forall_{y{\in}{\rm Dom}(Y)}\,P_{t}( Y{=}y\mid X{=}x)=P_{u}(Y{=}y\mid X{=}x).
\end{multline}

{\em Full-concept drift}, also referred to as {\em severe drift\/} \citep{OnlineEnsembleLearningConceptDrift}, involves the posterior class distribution changing for all types of object.
\begin{equation} \label{eq:fullconcept}
\forall_{x{\in}{\rm Dom}(X)}\,\exists_{y{\in}{\rm Dom}\,(Y)}P_{t}(Y{=}y\mid X{=}x)\neq P_{u}(Y{=}y\mid X{=}x ).
\end{equation}

\subsubsection{Covariate Drift}
\label{subsubsec:DistDrift}
{\em Covariate drift}, or {\em virtual concept drift\/} as it is commonly called in the literature, occurs when the distribution of non-class attributes, $P(X)$, changes over time \citep{ProblemOfConceptDrift,Cieslak2009,StreamingData+ConceptDrift+Imbalance,UnifyingView,SurveyConceptDriftAdaptation}. 
Take, for example, a business that uses socio-economic factors to make predictions about customers.  Over time the demographics of the customer base may change, leading to a change in the probability of each demographic factor.

Formally, covariate drift occurs between times $t$ and $u$ whenever
\begin{equation}
P_t(X)\neq P_u(X).\label{eq:covariate_drift}
\end{equation}

\cite{UnifyingView} define \emph{covariate shift} adding two further constraints to Eq.~\ref{eq:covariate_drift}:
\begin{enumerate}
\item that it only occurs when the class label is causally determined by the values of the covariates ($X$ variables), and
\item that it only occurs when $P_t(Y\mid X)= P_u(Y\mid X)$.
\end{enumerate}
However, the value of the first constraint is questionable, because, as argued above, it is likely that in many real world learning tasks the causal relationships are mixed and many learners  utilize only the correlations between the covariates and the class.

Should it be useful to distinguish situations in which the second constraint is satisfied, we propose the term \emph{pure covariate drift}, defined as:
\begin{equation}\label{eq:PureCovariateDrift}
P_t(X)\neq P_u(X) \wedge P_t(Y\mid X)= P_u(Y\mid X).
\end{equation}
This identifies a situation where the $X$ distribution changes, but the posterior class distribution remains unaffected.

It seems desirable to retain the more general definitions of class and covariate drift (Eqs.~\ref{eq:ClassDrift} and \ref{eq:covariate_drift}) that we propose in addition to the pure forms, as it seems to be both coherent and useful to be able to make a statement that a case of drift includes both class and covariate drift.

\subsubsection{Novel Class Appearance}

\textit{Novel class appearance} is a special case of concept drift in which a new class comes into existence \citep{Masud2011}.  We treat this as a situation where $P_t(Y{=}y)=0$ for the new class $y$ at time $t$ and  $P_u(Y{=}y)>0$ at subsequent time $u$. 
Take, for example, a business that predicts which option a user will select on a web page.  If a new option is added to the web page then a new class is introduced.

\subsection{Drift magnitude}

We can define {\em minor drift\/} between successive stable concepts as
\begin{equation}
D(E_a, S_{a{+}1})<\gamma
\end{equation}
and {\em major drift\/} as
\begin{equation}
D(E_a, S_{a{+}1})\geq\gamma.
\end{equation}

To give an example of how drift magnitude will affect how the drift should be handled, if there is abrupt minor drift then it is likely to be appropriate to retain a model that is accurate for concept $a$ and to just refine it as evidence is gathered about concept $a{+}1$.  In contrast, if there is abrupt major drift then it might be best to simply abandon the previous model and start afresh with the evidence about the nature of the new concept as it becomes available \citep{nguyen2012heterogeneous}.

\subsection{Drift Frequency}

{\em Drift frequency} refers to how often concept drifts occur over a defined period of time \citep{kuh1991learning,presenceofconceptdrift}.  A~high frequency indicates that new concept drifts start within a short amount of time from each other, whereas a low frequency indicates that there are long intervals between drifts.


We define drift frequency $F_{[t,u]}$  relative to a time interval $[t,u]$:
\begin{equation}
F_{[t,u]}=|\{w\mid t\leq S_w\leq u\}|. 
\end{equation}

For an example of drift frequency comparison, consider two people, Alice and Bob.  They are repeating the same set of exercises with rests between each set.  Alice is fitter than Bob.  As such, Alice can complete the set quicker than Bob, and Bob needs longer rest than Alice. If the transition of their vital signs between resting and exercising and vice versa were compared, the frequency of Alice's transitions would be higher than Bob's.

\subsection{Drift Duration}
\label{subsec:DSpeed}
The literature identifies several distinct categories of drift duration. \citep{ProblemOfConceptDrift,StreamingData+ConceptDrift+Imbalance,LearningUnderConceptDrift,ConceptDriftProcessMining,TrackingDriftTypes,OnlineEnsembleLearningConceptDrift,ConceptDriftScenarioReview,brzezinski2010mining}.

{\em Abrupt drift}, or {\em sudden drift} occurs when a stream with concept $a$ suddenly changes to concept $a{+}1$ \citep{ProblemOfConceptDrift,StreamingData+ConceptDrift+Imbalance,LearningUnderConceptDrift,ConceptDriftProcessMining,TrackingDriftTypes,OnlineEnsembleLearningConceptDrift,ConceptDriftScenarioReview,brzezinski2010mining,AUE}.
A real world example of abrupt drift could be a market crash. In a stock market stream, almost instantly, stock values will change and follow a pattern different to previously.

Given that $\delta$ is some natural number $>0$ that defines the maximum duration over which abrupt drift can occur, abrupt drift between concepts $a$ and $a+1$ can be defined as occurring when:
\begin{equation} \label{eq:AbruptDrift}
S_{a{+}1}-E_a\leq \delta.
\end{equation}
The value of the constant $\delta$ will depend on the context of the data stream, and may be different for different streams.

For completeness we also define the alternative to abrupt drift, {\em extended drift}, which occurs when
\begin{equation}\label{eq:ExtendedDrift}
S_{a{+}1}-E_a> \delta.
\end{equation}

A real world example of extended drift could be a recession.  Unlike a market crash, stock values at the beginning of a recession will slowly change over an extended period of time. Eventually, the changes in all of the stock values will follow a different pattern to before the recession started.

\subsection{Blip Drift}
\label{subsec:BlipDrift}
{\em Blip drift\/} is a special case of abrupt drift coupled with very short concept duration. In~blip drift, the blip concept replaces the dominant concept for a very short period of time \citep{ConceptDriftScenarioReview}. 

In the literature, there has been some debate as to whether blip drift should be adjusted for during training, or whether it should even be considered a type of drift \citep{brzezinski2010mining}.  As such, a number of systems have been designed to ignore blip drift.  Those that do detect the change in concept and adjust their models when a blip occurs may lose accuracy if they are unable to detect that the change is a blip and restore the old model when the blip concept has gone.

It is important to note that blip drift is different to what is defined as an {\em outlier}.  An~outlier is a single example from a stream which is anomalous in the context of the concept at the time the example occurs.  A blip is a short sequence of examples that are part of a single concept.  As such, outliers fall under the problem of anomaly detection, which is an entirely different topic.

A real world example of blip drift is the Cyber Monday sale. Cyber Monday is the first Monday after US Thanksgiving, where online stores persuade customers to shop online and offer significant discounts on their merchandise for that day only. If an online store had a stream that reported sales statistics at regular intervals, all samples from Cyber Monday would be significantly different to most other times in the year. 

Given that $\beta$ is the maximum duration for a blip drift to occur, concept $a$ can be defined as blip drift if
\begin{gather} \label{eq:BlipDrift}
S_{a+1} - E_{a} \leq \beta.
\end{gather}

The value of $\beta$ will depend on the context of the data stream, and may be different for every stream.

\subsection{Concept Transition}

There has been a wide diversity of terms and descriptions relating to the manner in which a transition between two concepts unfolds \citep{AUE,ProblemOfConceptDrift,LearningUnderConceptDrift,ConceptDriftProcessMining,TrackingDriftTypes,OnlineEnsembleLearningConceptDrift,ConceptDriftScenarioReview,StreamingData+ConceptDrift+Imbalance,brzezinski2010mining}.  This is another area in which the process of formalization has revealed a range of inter-related issues.

The first issue is whether the drift is a gradual progression of small changes or involves major abrupt changes. Gradual changes may or may not be a steady progression from one concept towards another. Fig.~\ref{fig:gradinc} illustrates some examples of how such changes might vary from a direct progression from one stable concept to another.  Given that $\mu$ is a maximum allowed difference between concepts over time period $\nu$  during a period of drift for the drift to be considered gradual, we define {\em gradual drift\/} between concepts $a$ and $a{+}1$ as:
\begin{equation} \label{eq:GradualDrift}
\forall_{t\in [E_a,S_{a{+}1}-\nu]}\, D(t,t+\nu)\leq \mu.
\end{equation}

A further issue is whether the change is a steady progression from concept $a$ toward concept $a{+}1$ such that at each time step the distance from concept $a$ increases and the distance to concept $a{+}1$ decreases.  We call this {\em incremental drift\/} and define it as follows:
\begin{multline} \label{eq:IncrementalDrift}
\forall_{t\in (E_a,\,S_{a{+}1})}\, \forall_{u \in (t,\,S_{a{+}1})}\, D(E_a,t)\leq D(E_a,u) \, \wedge\\ D(t,S_{a{+}1})\geq D(u,S_{a{+}1}).\end{multline}

An example of incremental drift is where credit card fraud patterns change over time.  Consider the introduction of RFID chips in credit cards.  The new technology changes the types of fraud that can be committed.  As more credit card customers get new cards with RFID chips, the new types of fraud become more common until everyone has new cards, where the concept drift in the transaction stream would stop.

Not all drift is incremental. {\em Probabilistic drift\/} \citep{OnlineEnsembleLearningConceptDrift} occurs when there are two alternating concepts such that one initially predominates and over time the other comes to predominate \citep{ProblemOfConceptDrift,LearningUnderConceptDrift,ConceptDriftProcessMining,TrackingDriftTypes,OnlineEnsembleLearningConceptDrift,ConceptDriftScenarioReview,StreamingData+ConceptDrift+Imbalance}. Consider a sensor network node.  Probabilistic drift would occur in a stream when a sensor network node is replaced.  A node cannot be immediately swapped; the new node must be tested to ensure it is working correctly.  As such, the two nodes will be turned on and off while the tests are conducted.  In terms of the sensor network stream, samples from two different concepts are flowing from that node, one from the faulty node and another from the new node.  The data from the new node will become more likely to occur in the stream, until only the concept from the new node is present.

Given that $f_{k}$ is a monotonically increasing function, indicating the probability of the new concept being in force, such that $f_{0}=0$ and $f_{S_{a{+}1}-E_a}=1$, probabilistic drift can be defined as:
\begin{equation}
\label{eq:ProbabilisticDrift}
\forall_{t\in [0, S_{a{+}1}-E_a]}\, \forall_{o}\, P_{E_a+t}\left({\mathcal X}{=}o\right)= 
\left( 1{-}f_{t} \right)P_{E_a}\left({\mathcal X}{=}o\right) +{f_{t}P}_{S_{a{+}1}}\left({\mathcal X}{=}o\right). 
\end{equation}
Depending on the nature of $f$, probabilistic drift may be gradual, and if so is likely to be incremental, but is not necessarily so.

Note that there has been some debate as to whether probabilistic and incremental drift are actually different. 
\cite{TrackingDriftTypes} and \cite{StreamingData+ConceptDrift+Imbalance} do not distinguish between the two types.  \cite{brzezinski2010mining} raises the issue that some of the literature ignores this difference.  Our formal definitions make it clear that they are in fact different, yet somewhat related, as expressed in \cite{brzezinski2010mining}.  Incremental drift may be probabilistic, but is not necessarily so, and probabilistic drift may be incremental, but is also not necessarily so.  To our knowledge we are the first to formally identify an alternative form of incremental drift to probabilistic drift.

\subsection{Drift Recurrence}
\label{subsec:DReoccur}
When drift occurs, the new concept may either be one that has not previously appeared in the data stream or may be a recurrence of a pre-existing concept. The latter is known as {\em drift recurrence}. There are a number of ways in which concepts can recur in a stream \citep{OnlineEnsembleLearningConceptDrift,ConceptDriftProcessMining,ConceptDriftScenarioReview,brzezinski2010mining,LearningUnderConceptDrift,StreamingData+ConceptDrift+Imbalance,LearningRecurringConcepts}.

A real world example of recurring drift could be in the use of a phone app.  A user using a particular app could use it in a certain way when they are at home compared to how they use it when they are at work.  The recurring concepts would be the use at home and the use at work.  These concepts of app use would recur whenever a user arrives at home or at work.

Recurring drift can be defined as:
\begin{equation}\label{eq:ReoccurDrift}
\exists_a\,\exists_b\, a\neq b \wedge D\left( S_a,S_b\right)= 0.
\end{equation}

\subsubsection{Cyclical Drift}
\label{subsubsec:cyclical}
As the name implies, {\em Cyclical Drift\/} is a form of recurring drift that occurs when two or more concepts recur in a specific order \citep{ProblemOfConceptDrift,StreamingData+ConceptDrift+Imbalance}. A good example of this is the weather patterns in a city. Assuming no climate change is present, meteorologists can expect the patterns in the weather to recur at particular times of the year when compared to previous years. The concepts could be defined as the four seasons, with incremental drift occurring between each season. This cycle in the drift would renew itself at the start/end of each year.

Given a stream has $i$ concepts in a cycle, cyclical drift can be defined as:

\begin{equation} \label{eq:cyclical}
\forall_a\,  D\left( S_a ,S_{a+i}\right)=0.
\end{equation}

\subsubsection{Cycle duration}

One issue that naturally arises in the context of recurring concept drift is the periodicity of the drift recurrence.
Our formal analysis of this issue has revealed that this actually encompasses multiple dimensions.  For example, winter may always occur at a particular time of year, but the exact start date and duration of winter may vary from year to year.  In contrast, the working day might always start and end at precisely defined times.

If a cycle contains $i$ concepts, the {\em duration\/} of a cycle starting at concept $a$ is
\begin{equation}
S_{a{+}i}-S_a.
\end{equation}

{\em Fixed frequency cyclical concept drift\/} occurs when a cycle has a fixed amount of time to occur.  For example, the cycle of the seasons must complete over the course of 365.24 days.  Fixed frequency cyclical concept drift of duration $m$ between $i$ concepts  can be defined as follows.
\begin{equation}
\forall_a\,  D\left( S_a ,S_{a+i}\right)=0 \wedge S_{a+i}\leq E_a{+}m \wedge E_{a+i}\geq S_a{+}m.
\end{equation}

A number of other aspects of concept periodicity might be fixed.

{\em Fixed concept duration cyclical  drift\/} occurs when every period of stability occurs for a fixed amount of time. 
For example, a store might conduct periodic one-day sales.  The relevant concepts might be shared between each sale period, which is always 24 hours. Fixed concept duration cyclical drift can be defined as:
\begin{equation} \label{eq:fixeddurationconcept}
\forall_a\,  D\left( S_a ,S_{a+i}\right)=0 \wedge E_a{-}S_a = E_{a{+}i}{-}S_{a{+}i}.
\end{equation}

{\em Fixed drift duration cyclical  drift\/} occurs when every period of drift occurs for a fixed amount of time.  For example, consider an athlete repeating a set of exercises with rest breaks between each set. The transition of his vital signs between resting and exercising and vice versa will have a fixed transition time. Fixed drift duration cyclical drift can be defined as:
\begin{equation} \label{eq:fixeddurationdrift}
\forall_a\,  D\left( S_a ,S_{a+i}\right)=0 \wedge S_{a{+}1}{-}E_a = S_{a{+}i{+1}}{-}E_{a{+}i}.
\end{equation}

{\em Fixed concept onset cyclical drift\/} occurs when the start of a period of stability begins at the same time in every cycle. 
For example, a store might open at the same time every day.  It will enter a state of being open but inactive and this state will last a differing period of time and drift to a fully active state in varying ways and at varying rates.
Fixed concept onset cyclical drift can be defined as:
\begin{equation} \label{eq:fixedonsetconcepts}
\forall_a\,\forall_b\,  D\left( S_a ,S_{a+i}\right)=0 \wedge S_{a+i}{-}S_{a} = S_{b+i}{-}S_b.
\end{equation}

{\em Fixed drift onset cyclical drift\/} occurs when the start of a period of drift begins at the same time in every cycle.
For example, a store might close at the same time every day.  At this time it will start a drift from an open and fully active state to a closed state.  The nature and duration of that drift will depend on the number of customers still in the store at closing time and the types of transactions that they are seeking to compete. 
Fixed drift onset cyclical drift can be defined as:
\begin{equation} \label{eq:fixedonsetdrift}
\forall_a\,\forall_b\,  D\left( S_a ,S_{a+i}\right)=0 \wedge  E_{a+i}{-}E_{a} = E_{b+i}{-}E_b.
\end{equation}

Varying frequency, concept duration, drift duration, concept onset and drift onset 
cyclical drift each can be defined by negation of the corresponding definitions above.


\subsection{Drift Predictability}\label{sec:predictability}

{\em Drift predictability}, as the name implies, describes how predictable some aspect of a drift is \citep{OnlineEnsembleLearningConceptDrift}.  Any aspect of drift might be more or less predictable, from when a drift starts, to when a drift ends, to what the subject of the drift is, and so on.  As this list of aspects of drift that could be predicted can be very long, we find that producing formal definitions for each type of drift predictability would be impractical.  Drift predictability is important however, as it should influence the selection of stream mining algorithms; affect the performance of algorithms that seek to anticipate drift; and inform the design of stream mining algorithms that will operate in environments with predictable drift, such as cyclical drift governed by seasonality.

Of the types of concept drift that we have defined, the only ones that are inherently predictable are fixed frequency, concept duration, drift duration, concept onset and drift onset concept drift.  These are predictable because there is a constant involved in each that might be inferred.  For example, drift due to time of day has a regular cycle and hence many aspects of it are relatively predictable.  In contrast, the onset, duration and magnitude of drift due to a stock market crash is relatively difficult to predict. 

\section{Case studies}\label{sec:CaseStudies}

We have asserted that our new quantitative characterizations of concept drift are critical to deep understanding of drift, its impacts on learning systems, the capacity of drift detection mechanisms to detect drift, the capacity of drift remediation mechanisms to adjust to drift and our capacity to design effective mechanisms for detecting and responding to drift.
In order to substantiate these assertions, we conduct a simple proof-of-concept pilot study, that provides significant insights that would not be possible without our quantitative measures.


\sloppy 
To this end, we generated two types of synthetic data stream containing \emph{abrupt} drift of a \emph{given magnitude}. The first type of stream was subjected to abrupt pure class drift (see Section \ref{subsubsec:ClassDrift}, Eq.\ \ref{eq:PureClassDrift}) and the second was subjected to abrupt pure covariate drift (see Section \ref{subsubsec:DistDrift}, Eq.\ \ref{eq:PureCovariateDrift}).  We conducted the experiments in the MOA \citep{bifet2010moa} workbench. We studied the responses to each type of drift of the following key learners from MOA: 
AccuracyUpdatedEnsemble \citep{AccuracyUpdatedEnsemble}, AccuracyWeightedEnsemble \citep{AccuracyWeightedEnsemble}, DriftDetectionMethodClassifier \citep{DDM}, DriftDetectionMethodClassifierEDDM \citep{EDDM}, HoeffdingAdaptiveTree \citep{HoeffdingAdaptiveTree},
HoeffdingOptionTree \citep{HoeffdingOptionTree},
HoeffdingTree \citep{HoeffdingTree}, LeveragingBag \citep{LeveragingBag}, Naive Bayes, 
OzaBag \citep{OzaBagBoost}, OzaBagAdwin,
OzaBoost \citep{OzaBagBoost}
and OzaBoostAdwin \citep{OzaBagBoost,ADWIN}.

We have released on GitHub the source code that we have developed for the generation of the data as described below (\url{http://dx.doi.org/10.5281/zenodo.35005}).

We list and motivate below the experimental settings that we chose in order to highlight the response of state-of-the-art classifiers to drift. 

\subsection{Model of the distribution from which the data is drawn}
\begin{figure}
	\centering
    \begin{tikzpicture}
      \node[draw,circle] (x1) at (-2,0) {$x_1$};
      \node[draw,circle] (x2) at (-1,0) {$x_2$};
      \node[draw,circle] (x3) at (0,0) {$x_3$};
      \node[draw,circle] (x4) at (1,0) {$x_4$};
      \node[draw,circle] (x5) at (2,0) {$x_5$};
      \node[draw,circle] (y) at (0,-1) {$y$};
      \draw[->] (x1) -- (y);
      \draw[->] (x2) -- (y);
      \draw[->] (x3) -- (y);
      \draw[->] (x4) -- (y);
      \draw[->] (x5) -- (y);        
    \end{tikzpicture}
    \caption{Bayesian network structure that is used to generate synthetic data}
    \label{fig:BN-synth}
\end{figure}
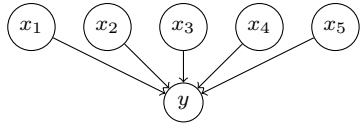

We generate categorical data from a Bayesian Network structure. 
The posterior distribution, $P(Y\mid X)$, is modeled by a full conditional probability table, which makes it possible to model any possible interaction between the different input variables. 
The structure of the network is given in Figure~\ref{fig:BN-synth}; we chose 5 parent nodes so that the posterior probability distribution is relatively difficult to learn. 
Each parent attribute takes 3 values and its multinomial probability is sampled following a flat Dirichlet with concentration parameter 1. We then assign, at random, a single class to each combination of $X$ values. To this end, the posterior probability $P(Y{=}y\mid X{=}\langle x_1,\cdots,x_5\rangle)$ is sampled independently for each combination $\vec{x}$ of $x_1,\cdots,x_5$; we start by choosing one of the 3 possible class values $v_c\sim\mathcal{U}(1,3)$ and assign $P(Y{=}v\mid X{=}\vec{x})=1\text{ if }v=v_c$, otherwise $0$. This allows us to know that the error of the Bayes optimal classifier is 0. Note that this tends to favor tree classifiers over linear classifiers such as Naive Bayes, as tree classifiers can build a precise model of the distribution. 

\subsection{Generation of the data}
We then generate a set of 100 synthetic data streams, each with 300,000 time steps ($t=1$ to $t=300,000$), and with one instance arriving at each time step. The stream is generated from two concepts, with an abrupt drift solely in the posterior probability at $t=100,000$. 
We then study the behavior of the classifiers over the following time steps $t=100,001$ to $t=300,000$. 

\subsubsection{Generating abrupt pure class drift with given magnitude}
For each data stream that we generate, we start by generating a random distribution over the covariates, $P(X)$, following the setting described above. We then need two posterior probability distributions: one before the drift, $P_{1, \ldots 100,000}(Y\mid X)$, and one after, $P_{100,001, \ldots 300,000}(Y\mid X)$. The distribution before the drift is also generated following the process described above; we detail now how to generate the posterior distribution after the drift, \emph{i.e.} a posterior distribution with a given magnitude drift relative to the initial one. 

We use the Hellinger distance \citep{HellingerDistance} as a measure of the drift magnitude in the posterior. We choose the Hellinger distance in this paper because
\begin{enumerate}
\item most readers are familiar with the Euclidean distance, which is very similar to the Hellinger distance;
\item because of its nice mathematical properties including symmetry and taking values in $[0,1]$, and
\item we will see below that it is simple to generate a posterior probability with given magnitude distance from another one. 
\end{enumerate}
The Hellinger distance for our model follows: 
\begin{equation}
H^2(p,p') = \frac{1}{C}\sum_{v_1}\cdots\sum_{v_5}\frac{1}{\sqrt{2}}\left(\sum_{y}\left(p'(y|v_1,\cdots,v_5)^{\frac{1}{2}}-p(y|v_1,\cdots,v_5)^{\frac{1}{2}}\right)^2\right)^{\frac{1}{2}}
\end{equation}
where $C$ is the number of combinations of $x_1,\cdots,x_5$, \textit{i.e.} $3^5$ in our case. 

As we want to keep the Dirach form for the posterior for each combination of the parent attributes, there are only two choices that can be made for each combination of $x_1,\cdots,x_5$: either we keep $P(Y{=}y\mid X{=}\langle x_1,\cdots,x_5\rangle)$ unchanged, or we modify it by changing the class for which $P(Y{=}y\mid X{=}\langle x_1,\cdots,x_5\rangle)=1$. When unchanged, the partial Hellinger distance is 0 for this particular combination of values; it is 1 otherwise ($\frac{\sqrt{2}}{\sqrt{2}}$). We then have:
\begin{equation}
H^2(p,p') = \frac{k}{C} \, \Leftrightarrow \, k = H^2(p',p)\cdot C
\end{equation}
where $k$ is the number of combinations of $x_1,\cdots,x_5$ for which the posterior distribution is changed. For a given (desired) magnitude $H^2(p,p')$, we thus simply modify the posterior distribution for $k$ combinations of $x_1,\cdots,x_5$. The $k$ combinations are then chosen at random following $\mathcal{U}(1,\binom{C}{k})$. For each of these combinations, we then choose another class value, following the process described above, but ensuring that a different class value is chosen. 

\subsubsection{Generating abrupt pure covariate drift with given magnitude}
For each data stream that we generate, we start by generating a random posterior probability distribution $P(Y\mid X)$ following the setting described above. We then need two probability distributions over the covariates: one before the drift, $P_{1,\ldots 100,000}(X)$, and one after, $P_{100,000,\ldots 300,000}(X)$. The distribution before the drift is generated randomly following the process described above; we detail now how to generate the covariate distribution after the drift, \emph{i.e.} a covariate distribution with a drift of a given magnitude with respect to $P_{1,\ldots 100,000}(X)$. 

We use again the Hellinger distance \citep{HellingerDistance} as a measure of the drift magnitude which is defined as follows for covariates:
\begin{equation}
H^2(p,p') = \frac{1}{\sqrt{2}}\left(
	\sum_{v_1}\cdots\sum_{v_5}\left(p'(v_1,\cdots,v_5)-p(v_1,\cdots,v_5) \right)^2
\right)^{\frac{1}{2}}
\end{equation}
There is unfortunately no closed-form solution to determine the parameters of $p'$ from $p$ and a given magnitude. We thus resort to sampling to find the parameters of $p_{x_i,1\leqslant i\leqslant 5}$ that achieve the desired magnitude $H^2(p',p)$.

\subsection{Results}

This learning task favors decision trees.  It is not possible to fully describe the concepts before or after drift other than by producing a map from each attribute-value combination to a class, except perhaps in a few cases where by chance multiple combinations differing only in the values of a few attributes all map to the same class by chance.  As a result of this advantage to one particular type of model, we do not believe much should be concluded from the relative errors of the alternative classifiers when they have reached their asymptote at the end of the initial concept period, at $t=100,000$.

We do, however, believe that the responses to the abrupt drifts are revealing.

\subsubsection{Results for pure class drift}
\begin{figure}
\def\addfig#1{\hspace*{-6pt}\includegraphics[width=.5\textwidth]{#1.pdf}}
\begin{tabular}{cc}
\addfig{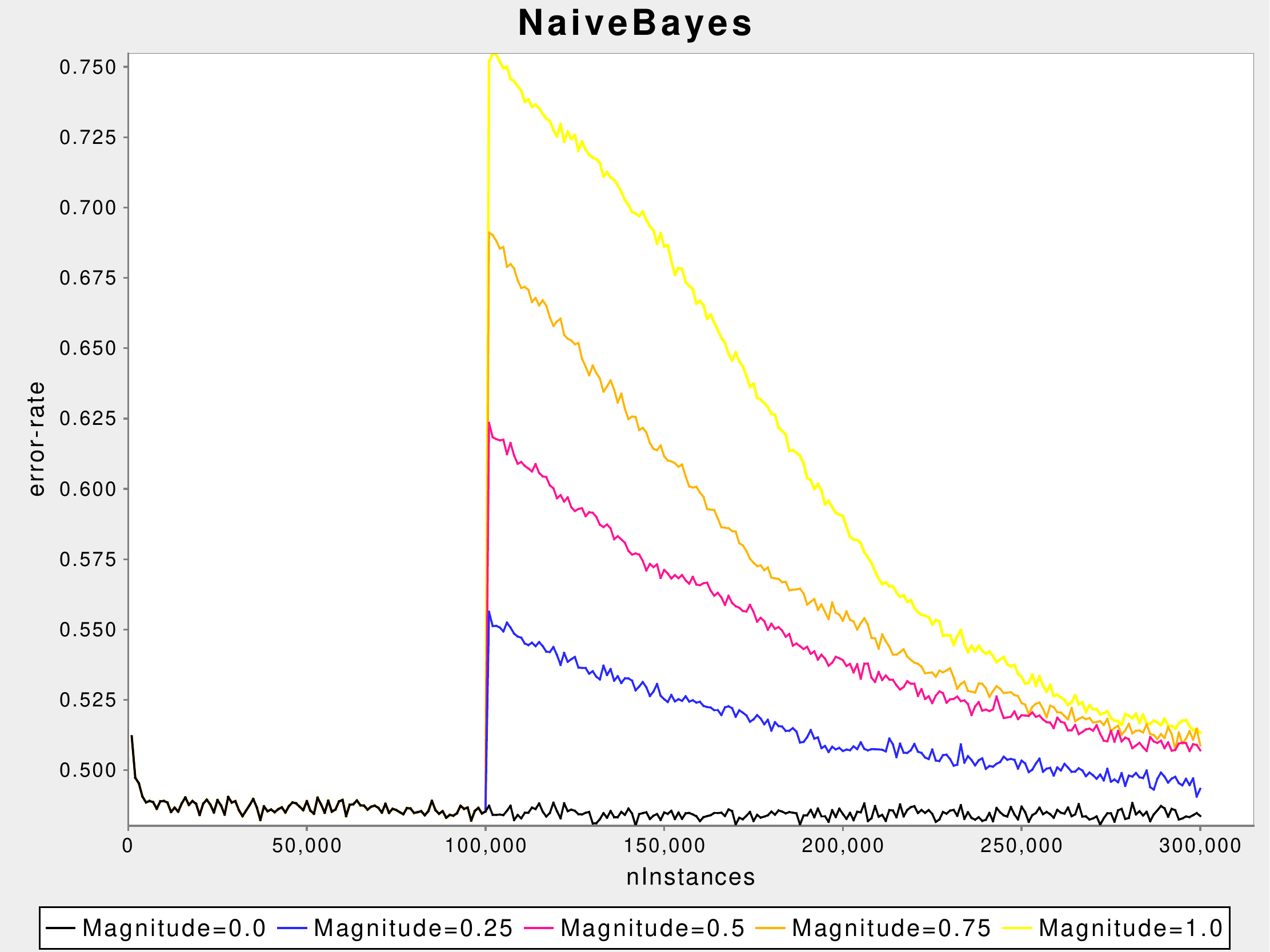}&
\addfig{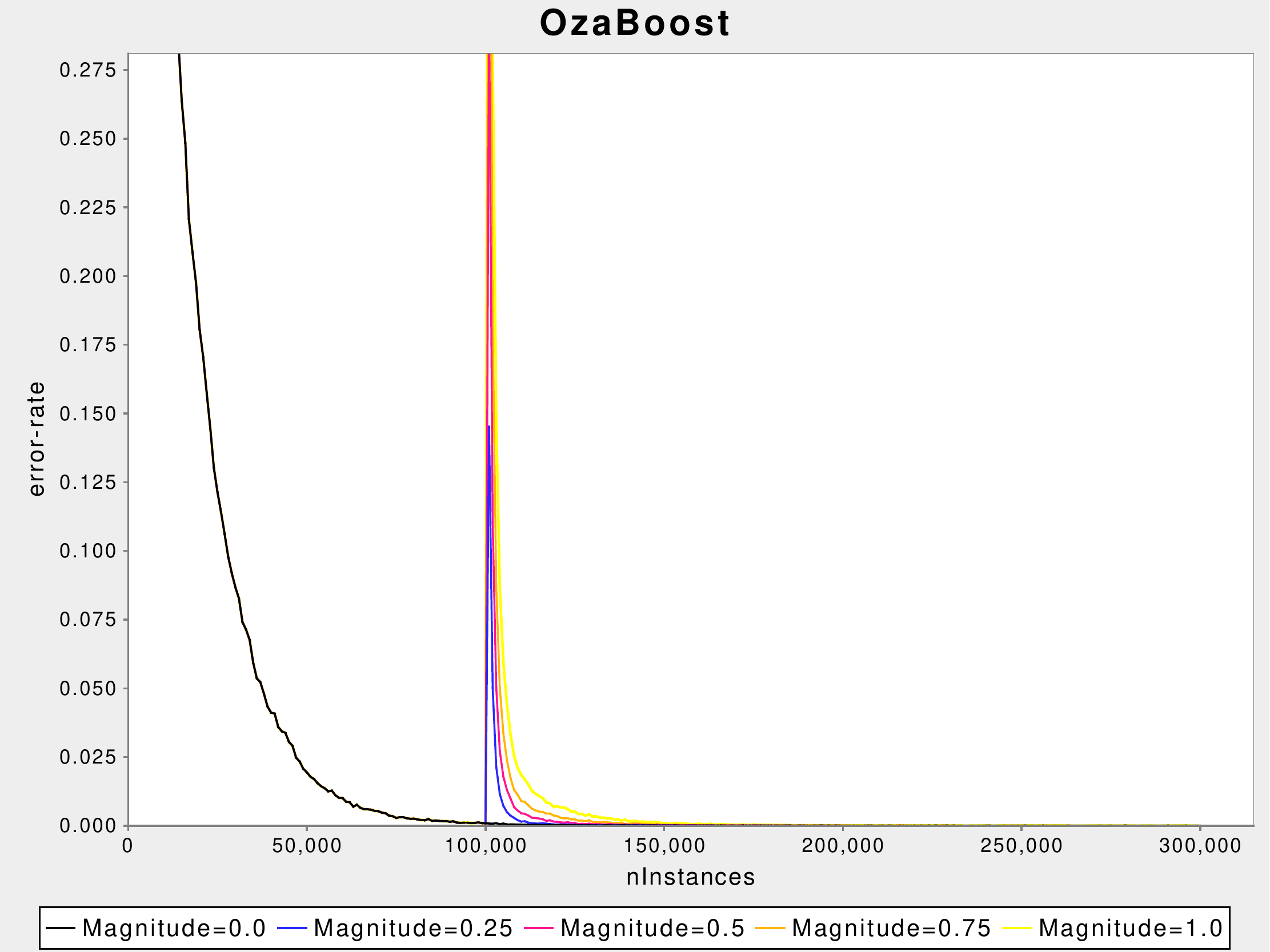}
\end{tabular}
\caption{Pure class drift: Learning curves with recovery times ordered by class drift magnitude}\label{fig:res1}
\end{figure}

There appear to be three classes of response.  The first is observed for Naive Bayes, which does not have any drift detection or remediation mechanisms, and OzaBoost.  This is perhaps the type of relative response to different magnitudes of drift that one would naively expect from most systems.  The greater the magnitude of drift the greater the immediate jump in error and the longer the time taken to recover the original error level. The learning curves for these algorithms are shown in Fig.~\ref{fig:res1}.
This and the following learning curves plot averages over each 1,000 times steps of the 0-1 loss for all of the 100 synthetic data streams --- thus each point plotted is an average of 100,000 values. 

\begin{figure}
\def\addfig#1{\hspace*{-6pt}\includegraphics[width=.5\textwidth]{#1.pdf}}
\begin{tabular}{cc}
\addfig{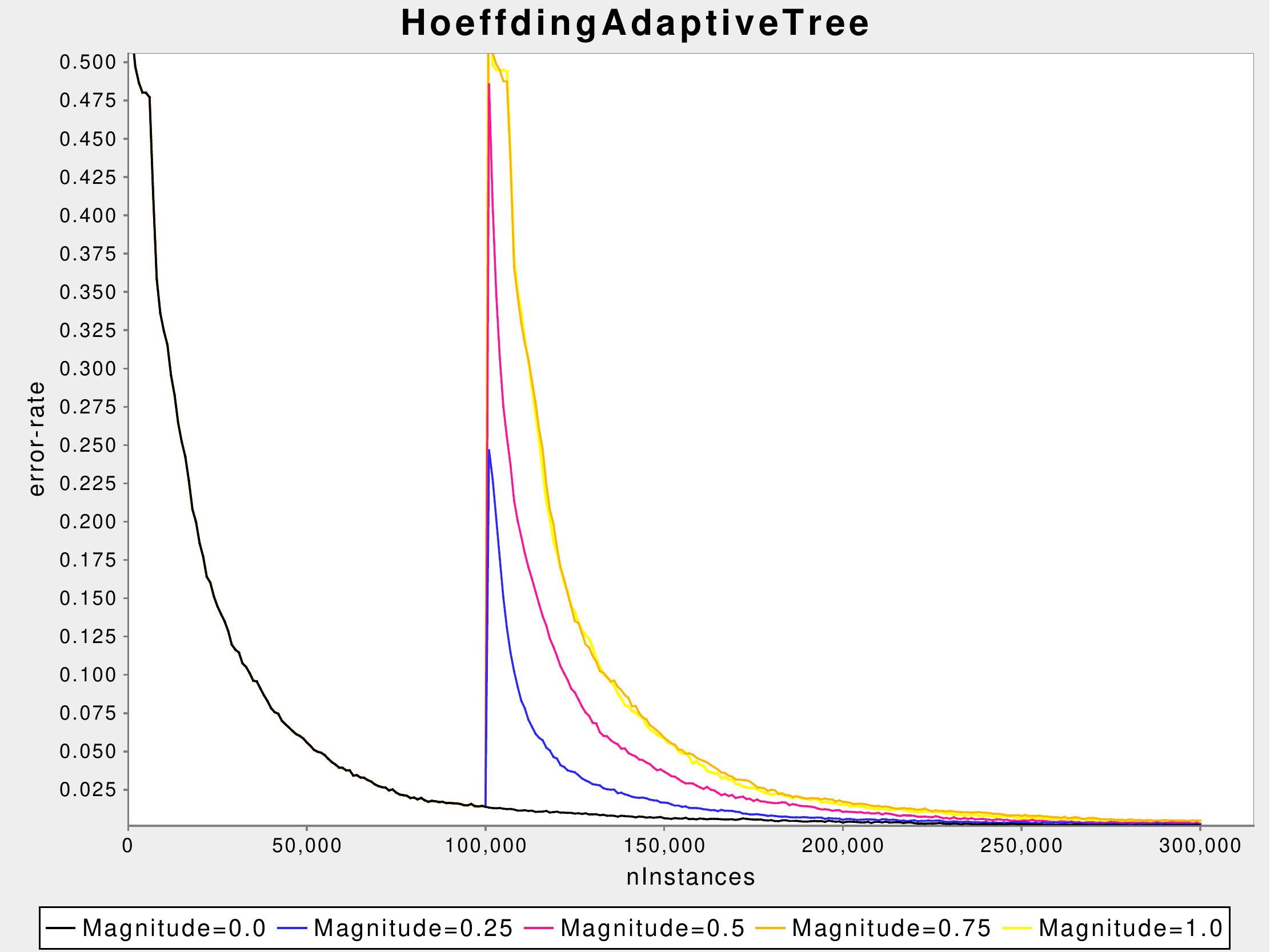}&
\addfig{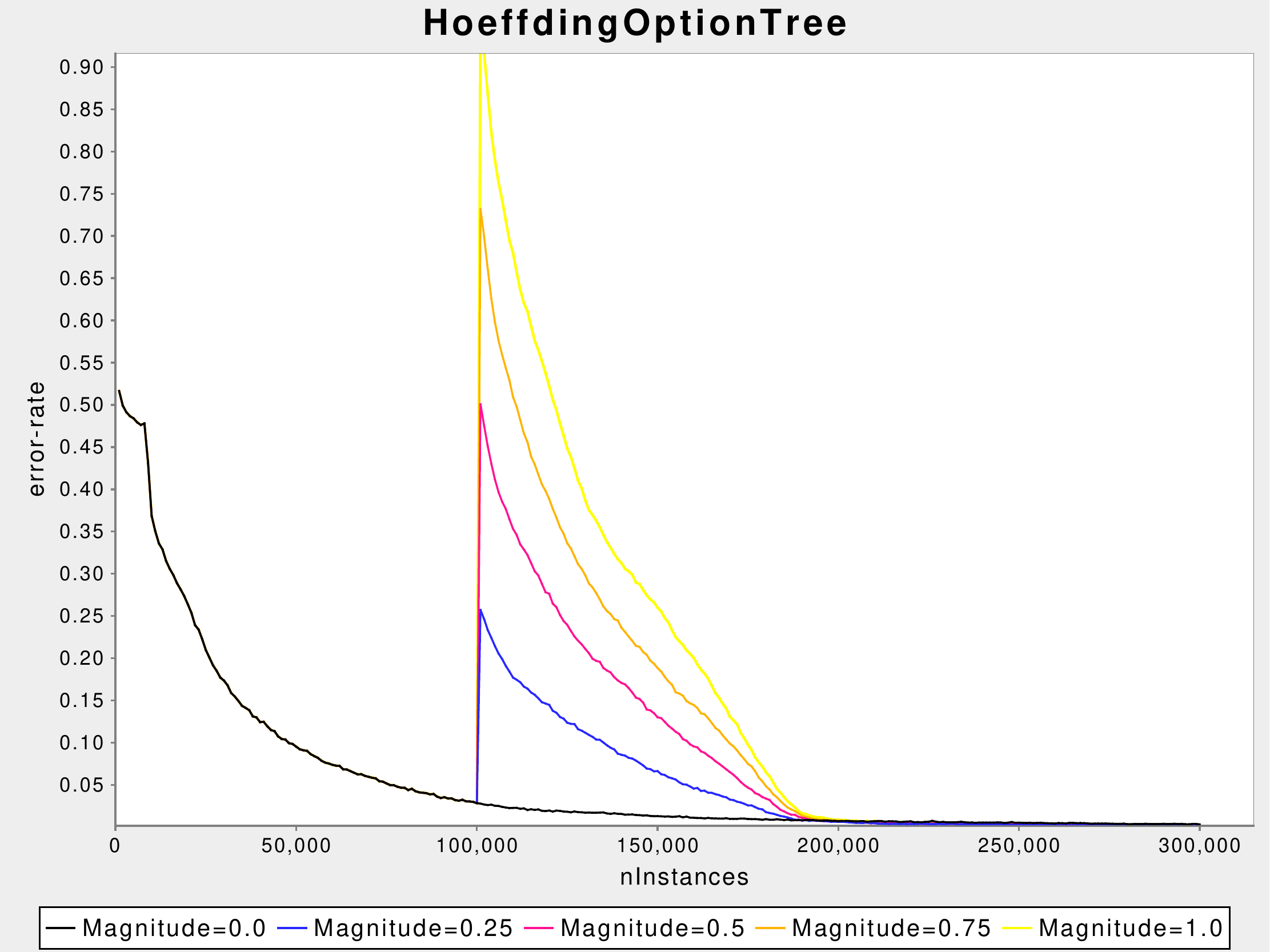}\\
\addfig{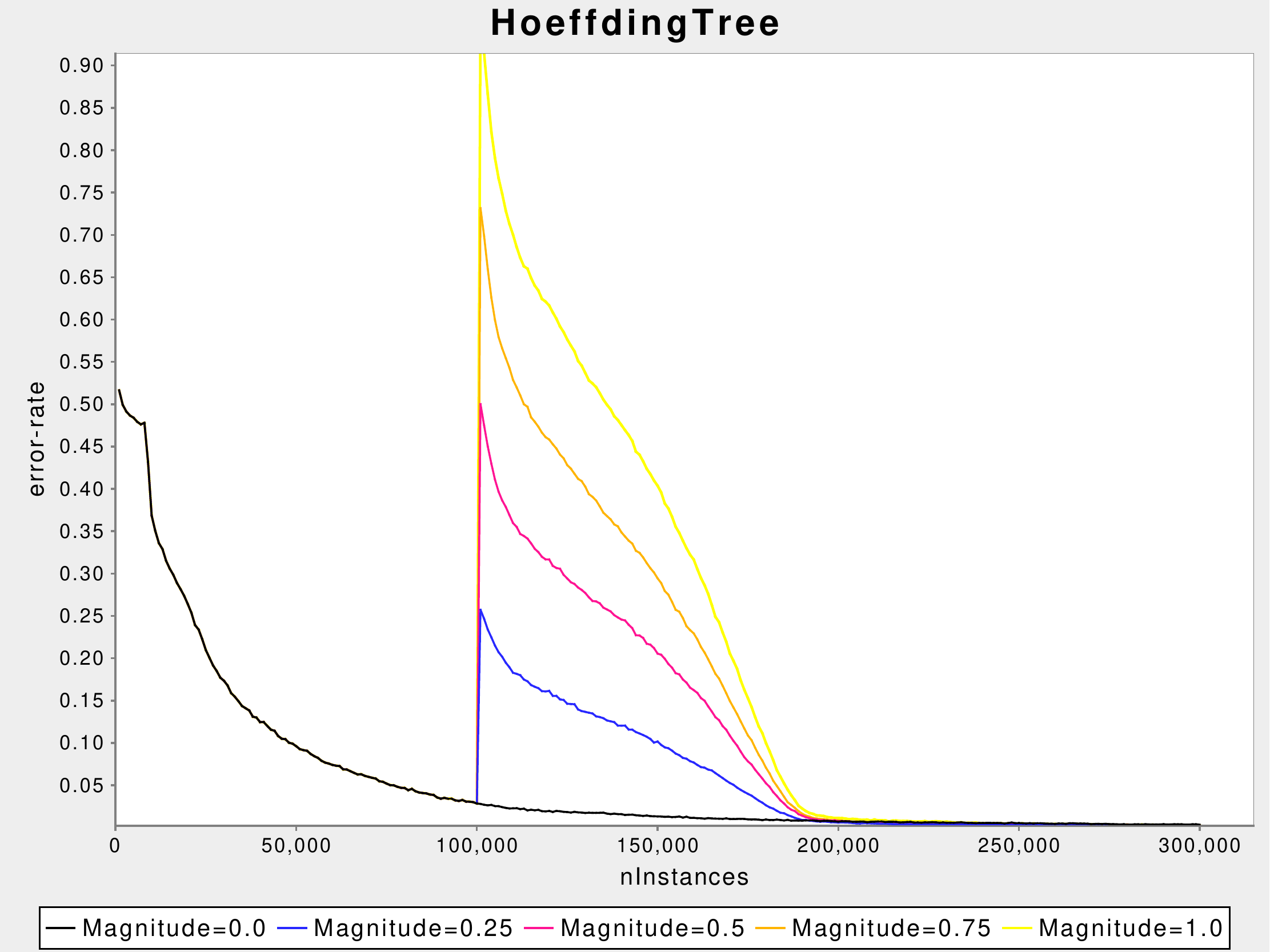}&
\addfig{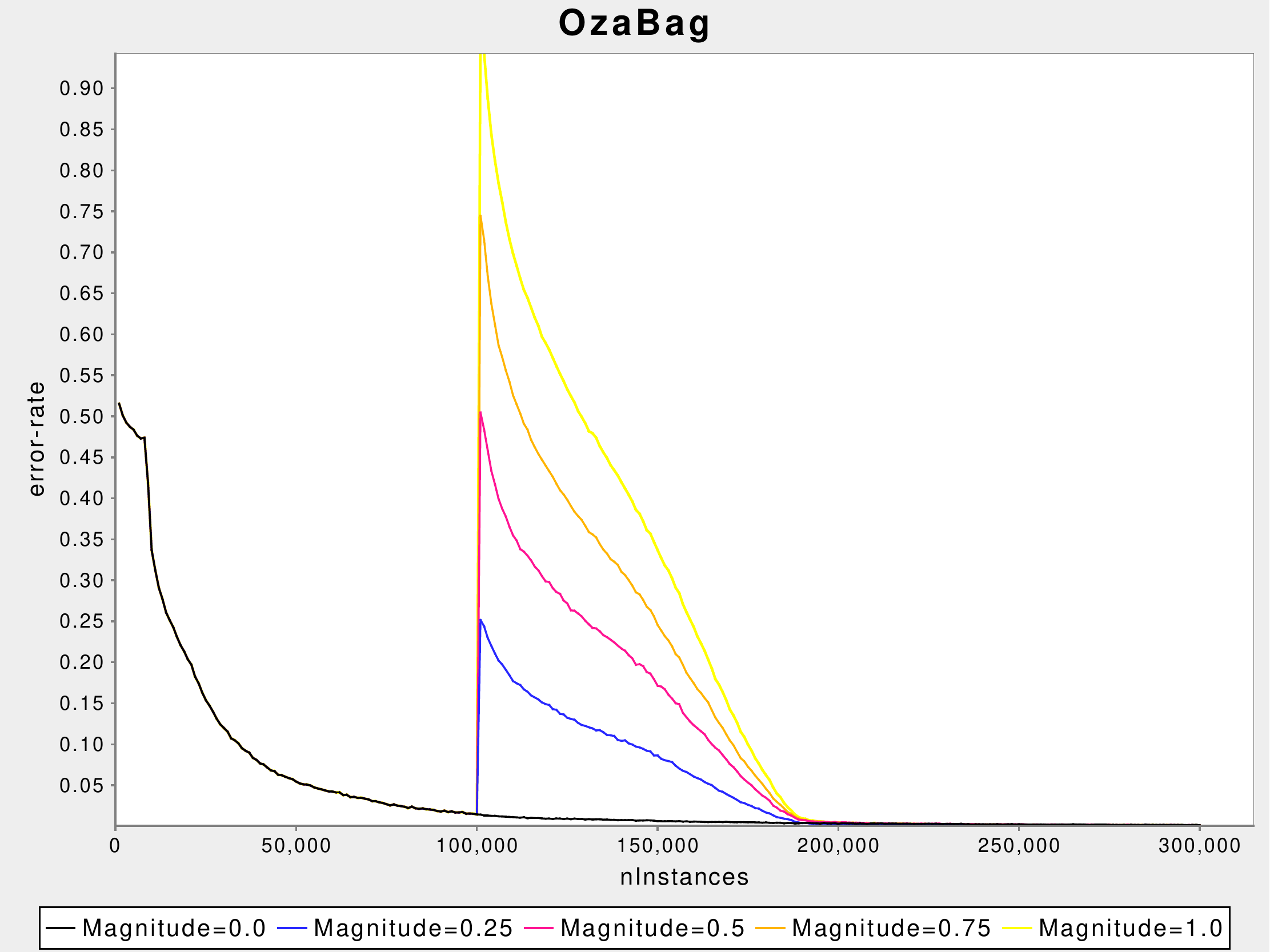}
\end{tabular}
\caption{Pure class drift: Learning curves with recovery time independent of class drift magnitude}\label{fig:res2}
\end{figure}

The second class of response is shared by HoeffdingAdaptiveTree, HoeffdingOptionTree, HoeffdingTree and OzaBag.  The greater the magnitude of the drift the greater the immediate jump in error, but all magnitudes of drift take the same length of time to recover back to the level of error without drift.  The learning curves for these algorithms are shown in Fig.~\ref{fig:res2}.

The third class of response is shared by AccuracyUpdatedEnsemble, AccuracyWeightedEnsemble, DriftDetectionMethodClassifier, DriftDetectionMethodClassifierEDDM, LeveragingBag, OzaBagAdwin and OzaBoostAdwin.  While the magnitude of the initial jump in error is ordered relative to the magnitude of the drift, rates of recovery are not.  For example, for DriftDetectionMethodClassifierEDDM magnitude 1.0 drift recovers much faster than magnitudes 0.75 or 0.5. \begin{figure}
\def\addfig#1{\hspace*{-6pt}\includegraphics[width=.5\textwidth]{#1.pdf}}
\begin{tabular}{cc}
\addfig{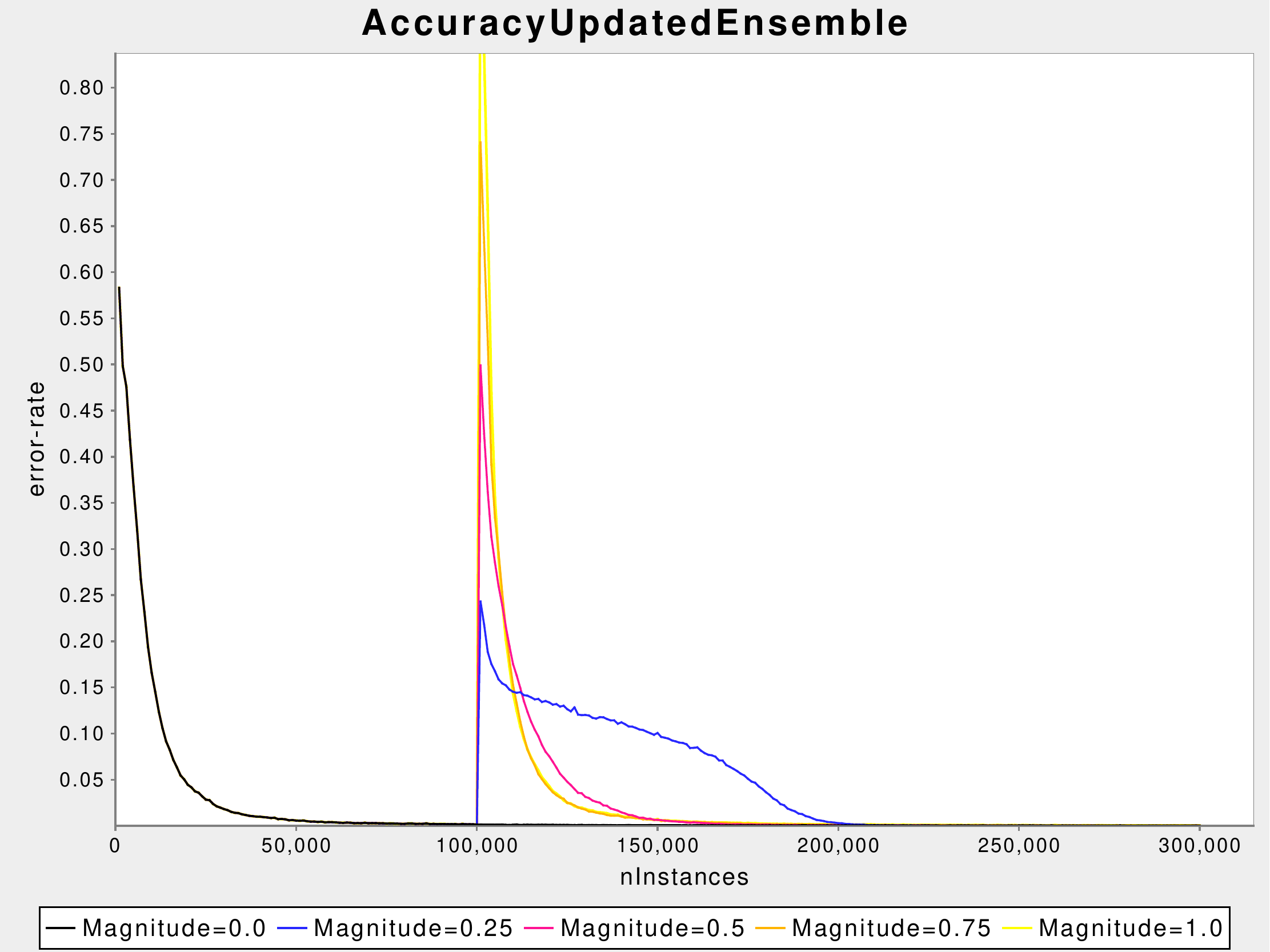}&
\addfig{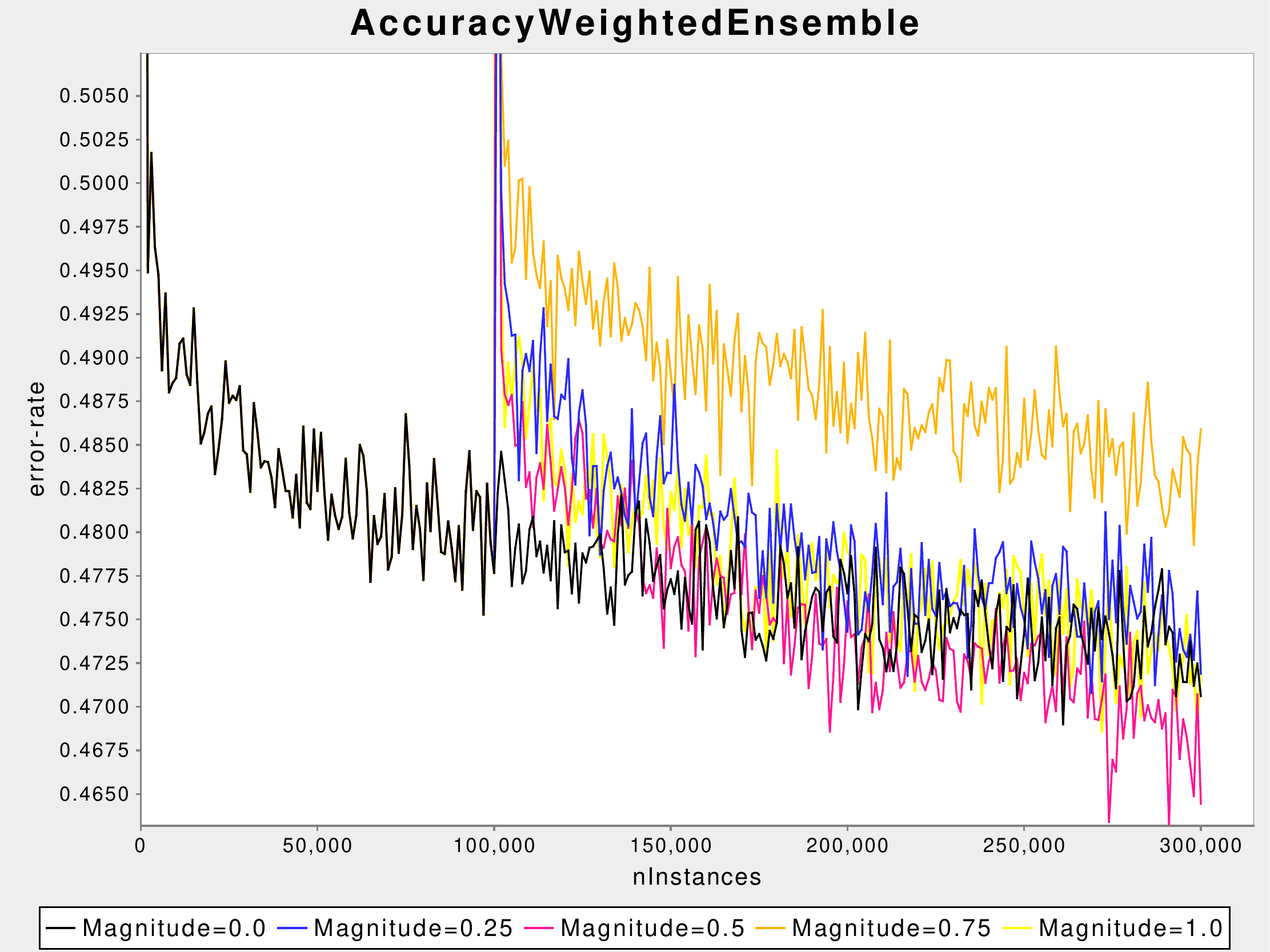}\\
\addfig{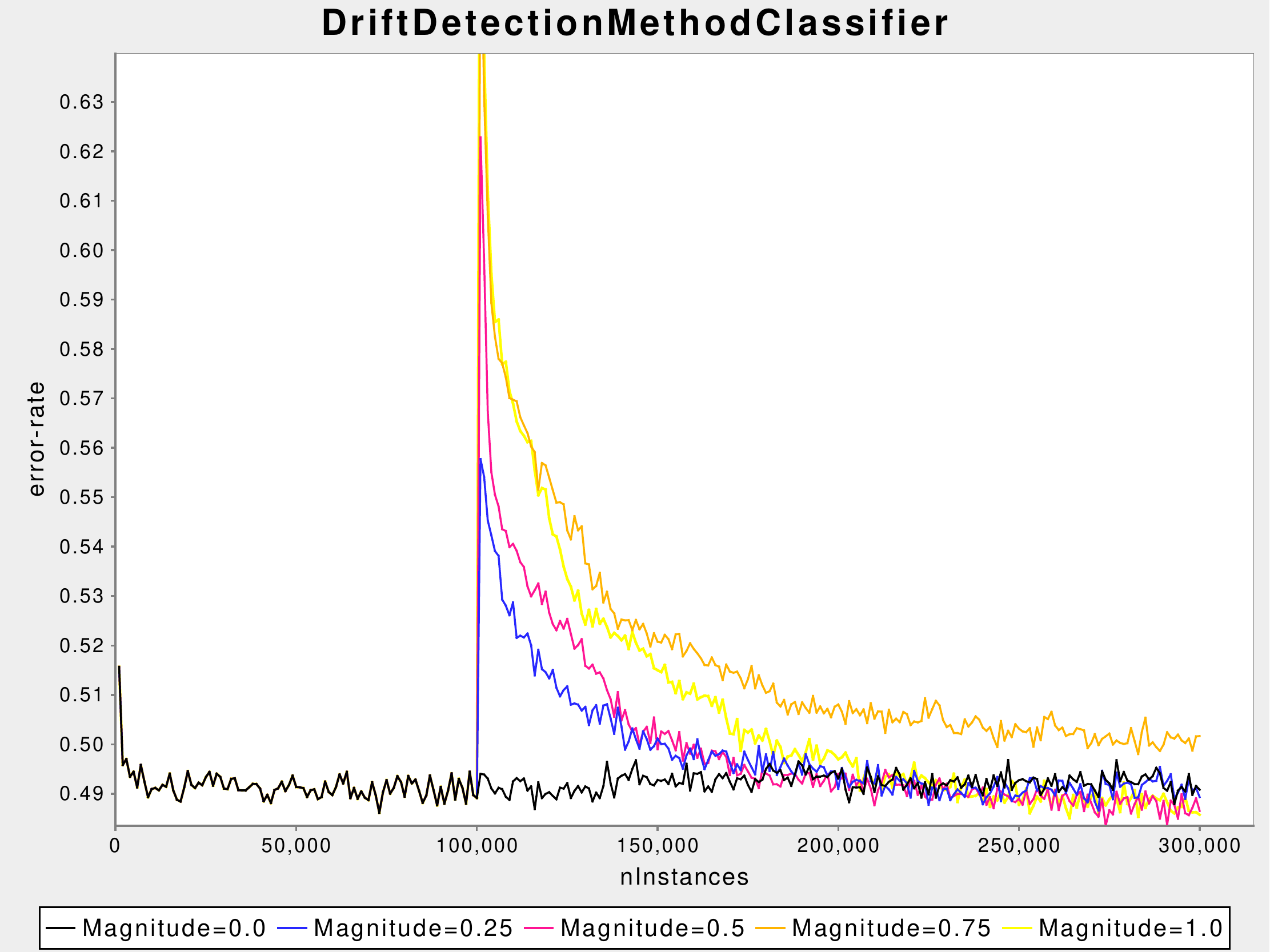}&
\addfig{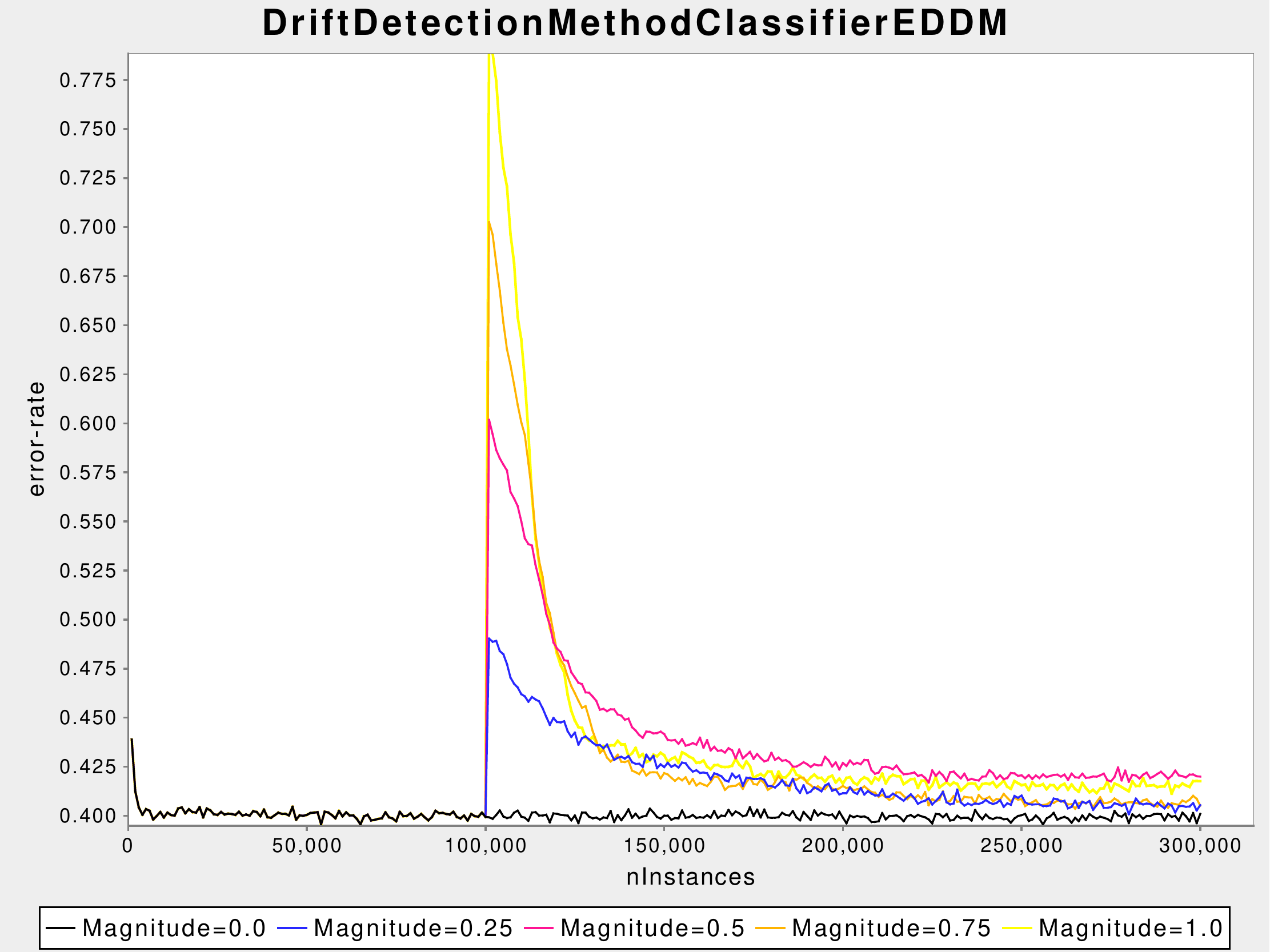}\\
\addfig{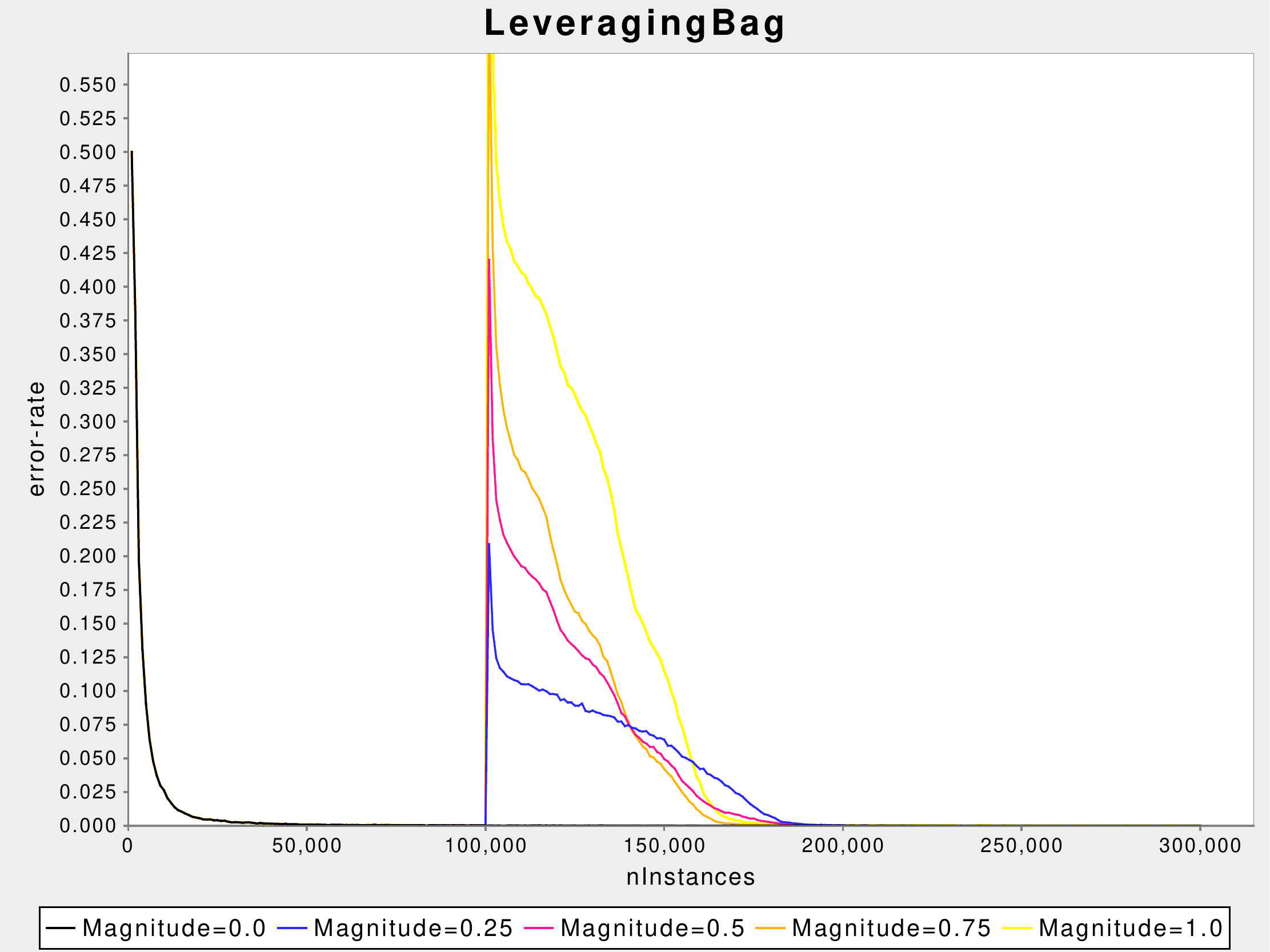}&
\addfig{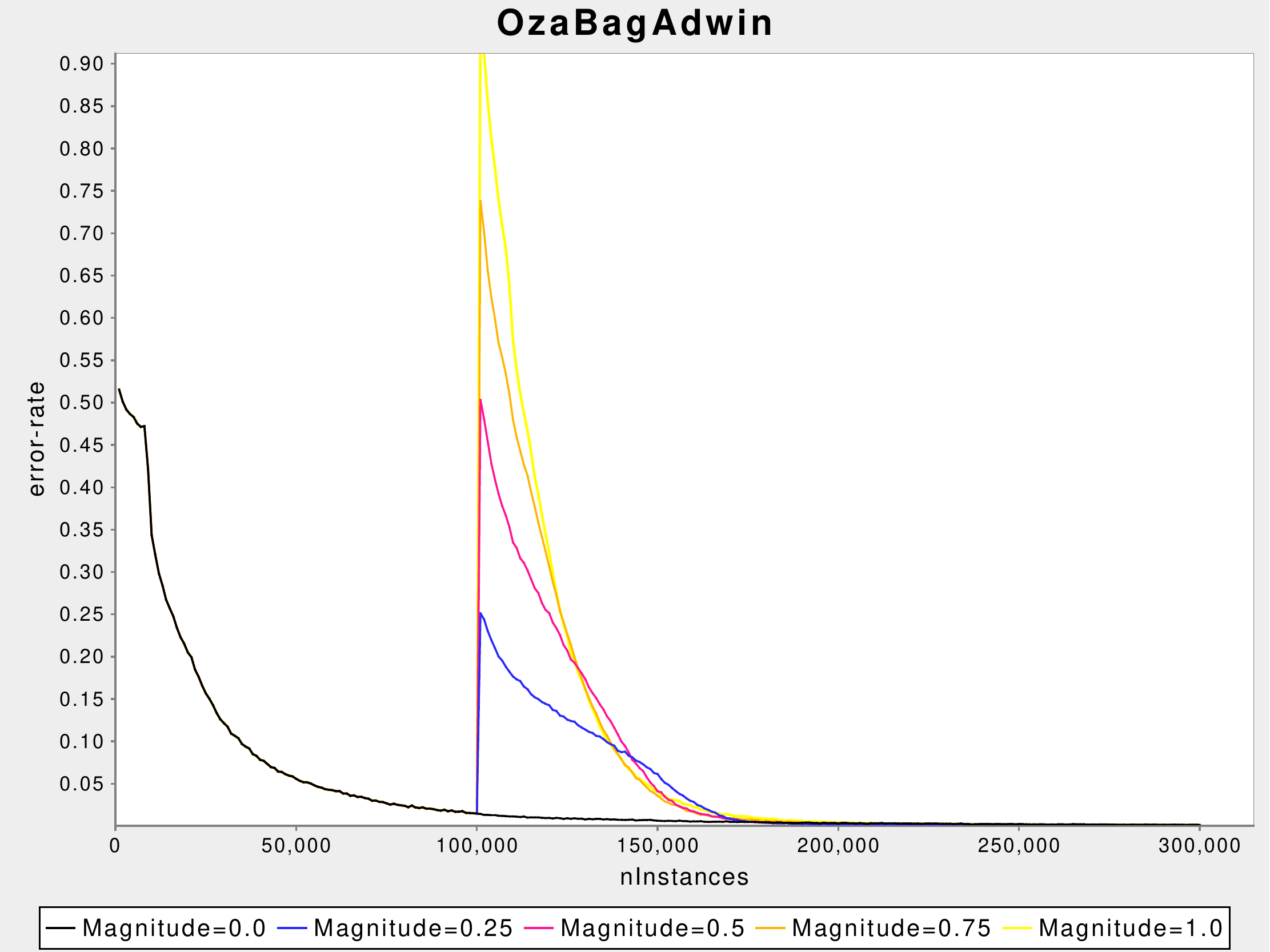}\\
\addfig{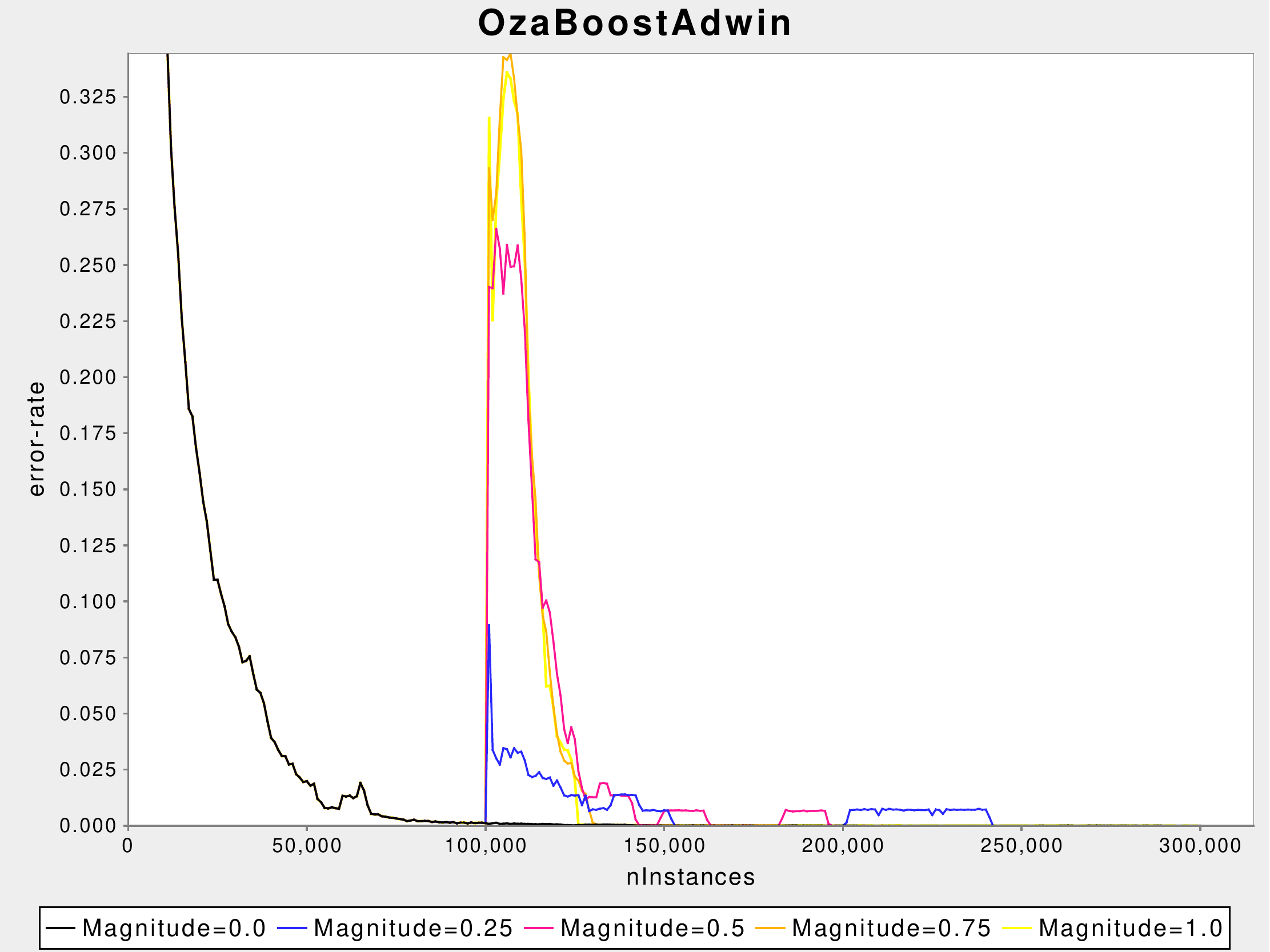}
\end{tabular}
\caption{Pure class drift: Learning curves with recovery time influenced but not ordered with respect to class drift magnitude}\label{fig:res3}
\end{figure}%
The learning curves for these algorithms are shown in Fig.~\ref{fig:res3}.

\subsubsection{Results for pure covariate drift}
A naive expectation for pure covariate drift might be that incremental discriminative classifiers should be unaffected by it while incremental generative classifiers should be more greatly affected as they utilize the joint distribution for classification and this changes even though the posterior does not.  There are some examples of these two expected patterns of response, but also some responses that differ greatly from that expected.  

\begin{figure}
\def\addfigcv#1{\hspace*{-6pt}\includegraphics[width=.5\textwidth]{CV#1.pdf}}
\begin{tabular}{cc}
\addfigcv{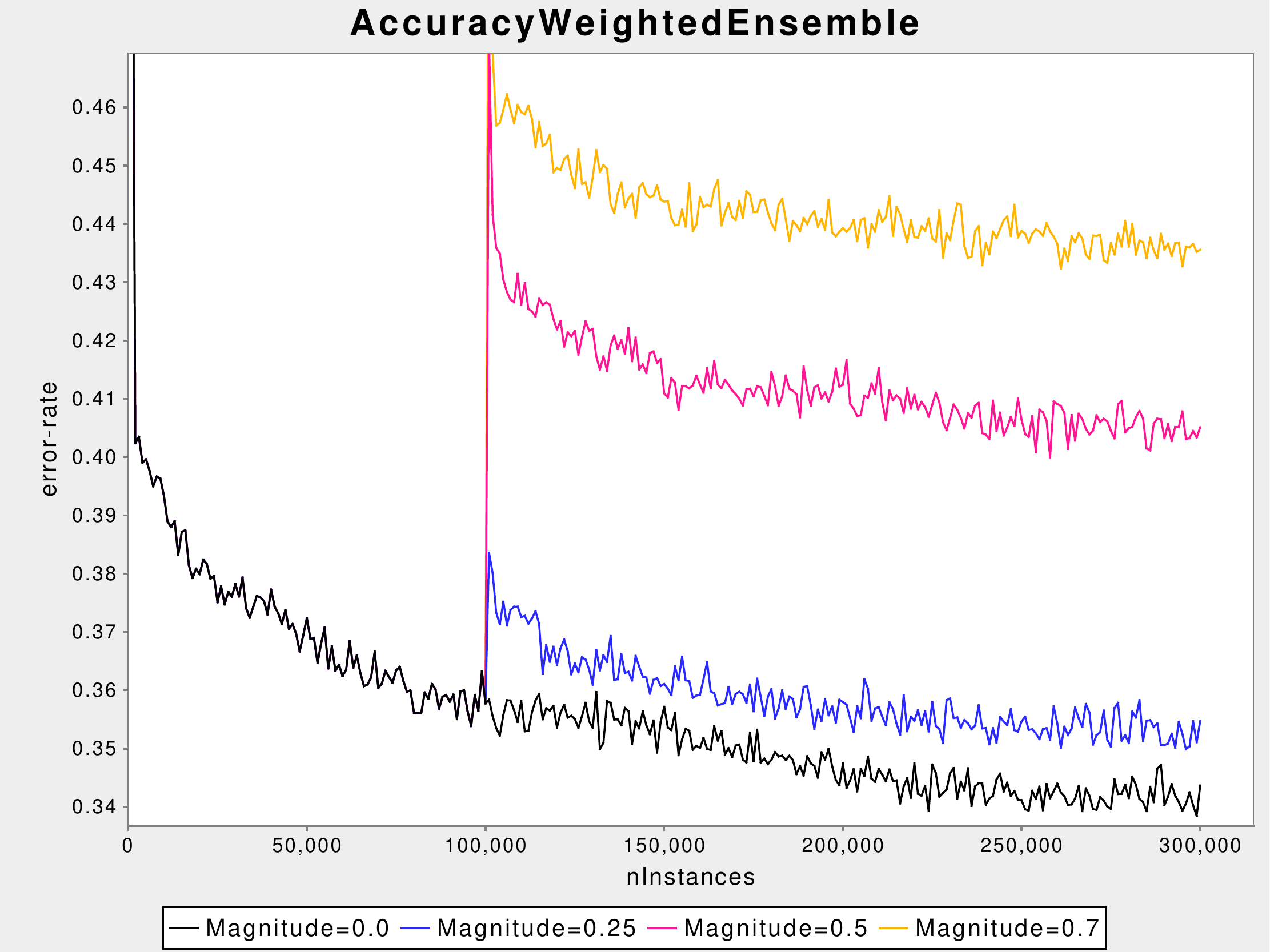}&
\addfigcv{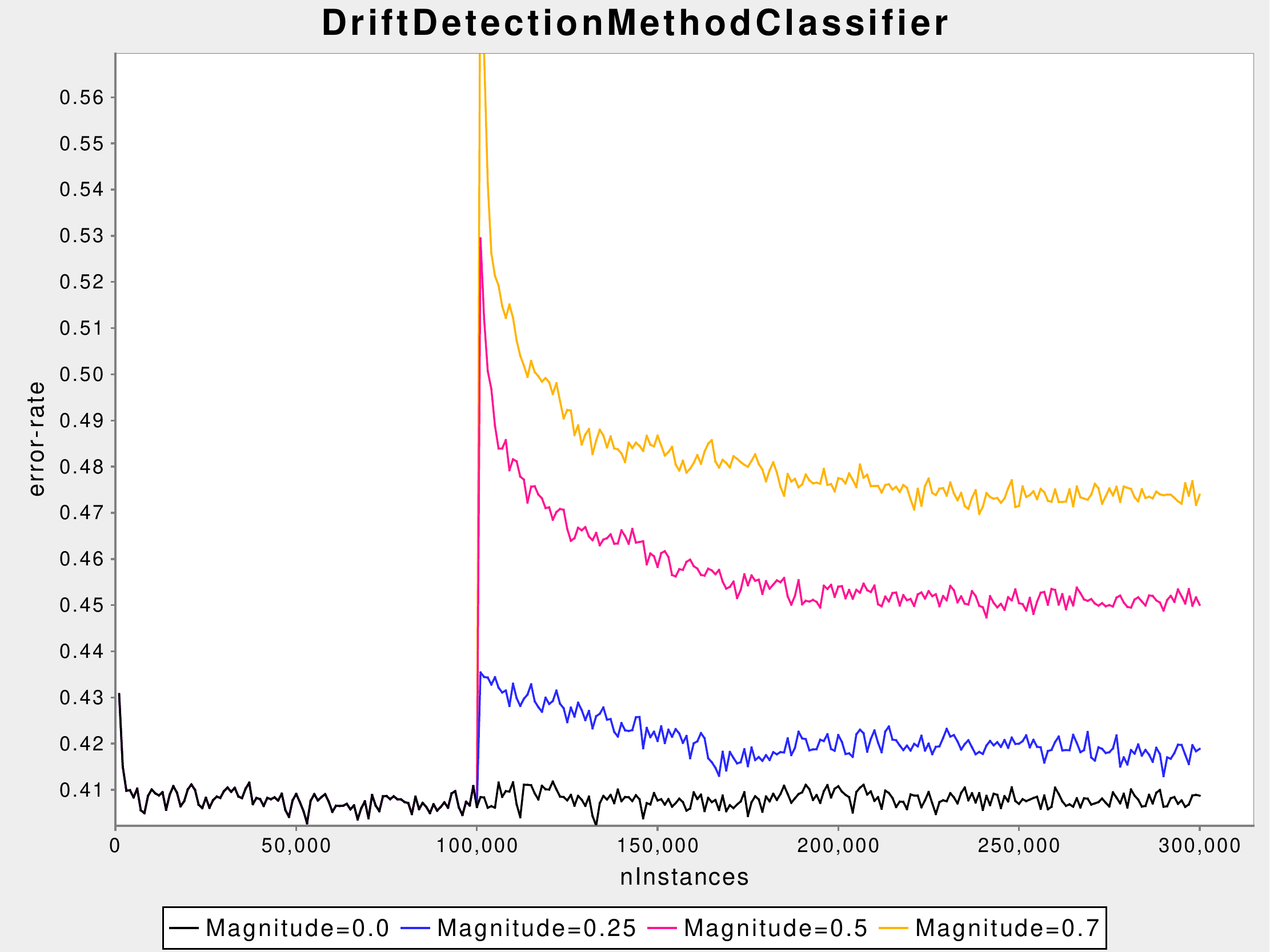}\\
\addfigcv{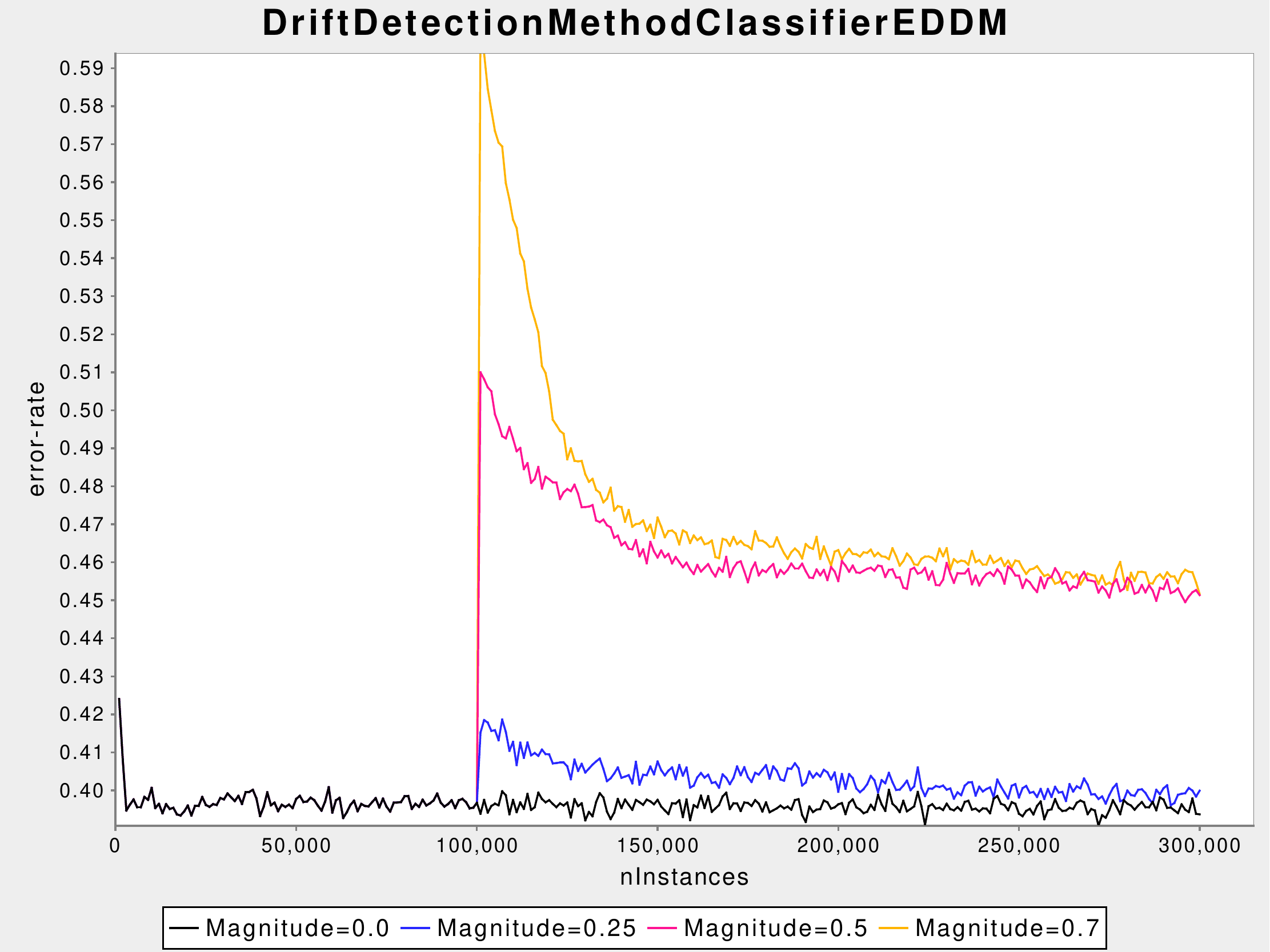}&
\addfigcv{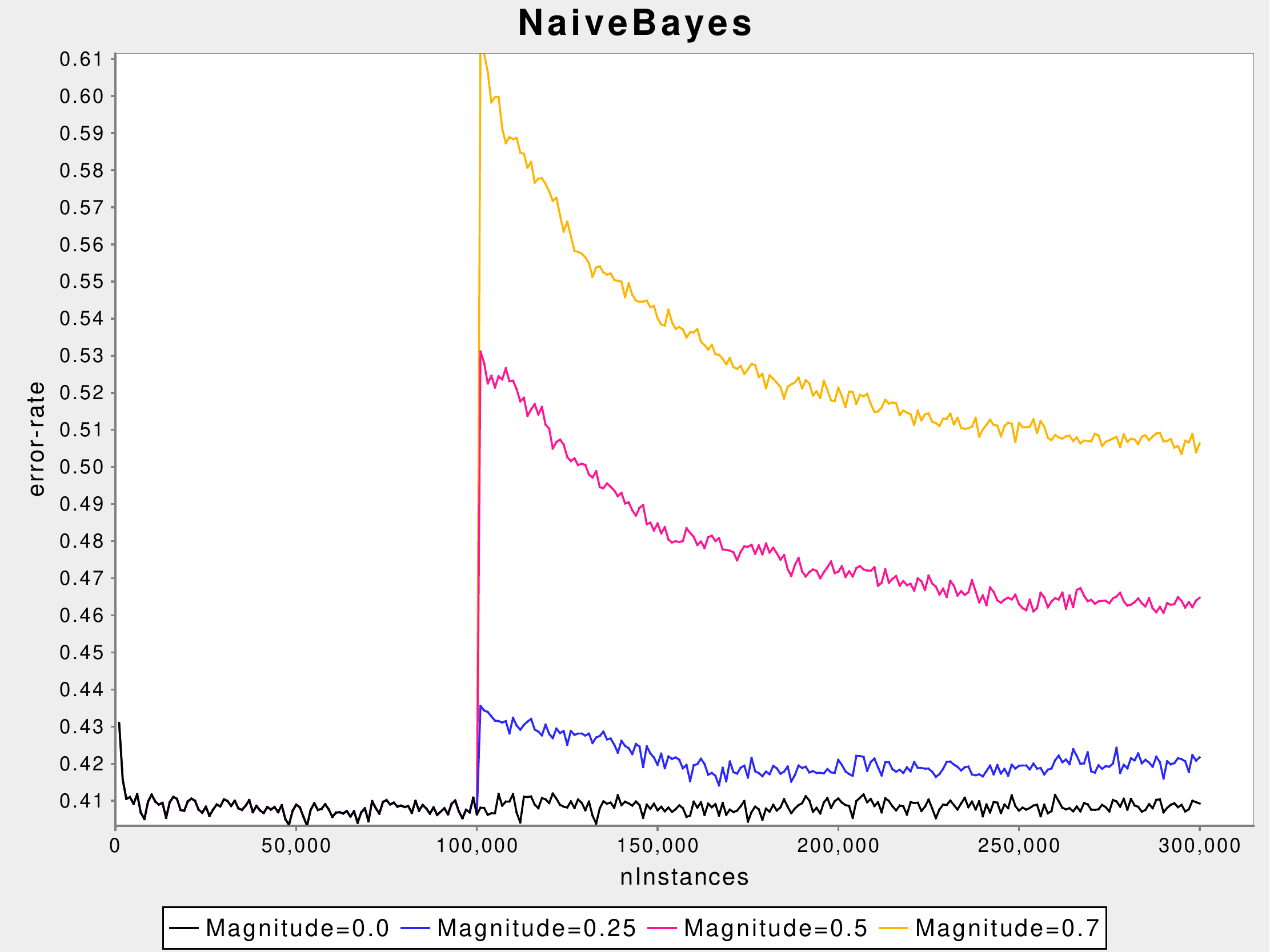}
\end{tabular}
\caption{Pure covariate drift: Learning curves in which the classifier does not recover}\label{fig:cvres1}
\end{figure}

In the first pattern of response the magnitude of the immediate jump in error and rate of recovery is ordered by drift magnitude, and the classifier does not recover back to the error obtained without drift.  
This is observed for all of the approaches that use Naive Bayes as the base classifier, regardless of the detection method that is used (\verb?NaiveBayes?, \verb?DriftDetectionMethodClassifer? and \verb?DriftDetectionMethodClassiferEDDM?). Although this pattern was anticipated with vanilla Naive Bayes, it seems that the drift detection methods only provide minor benefit in helping Naive Bayes to recover from the pure covariate drift. It is also extremely surprising to observe this phenomenon for tree-based discriminative classifier  \verb?AccuracyWeightedEnsemble?. The learning curves for these algorithms are shown in Fig.~\ref{fig:cvres1}.

\begin{figure}
\def\addfigcv#1{\hspace*{-6pt}\includegraphics[width=.5\textwidth]{CV#1.pdf}}
\begin{center}
\addfigcv{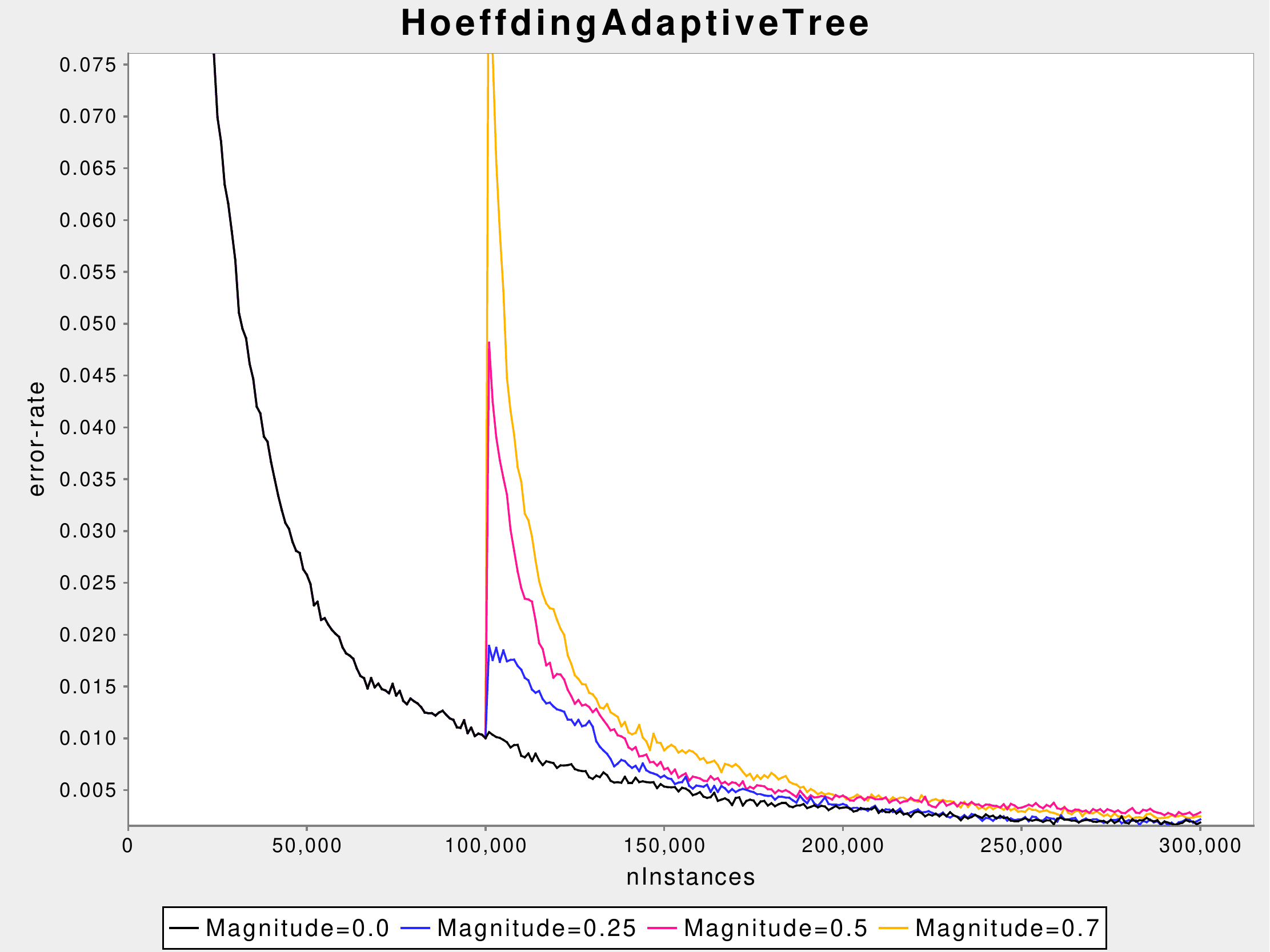}
\end{center}
\caption{Pure covariate drift: Learning curve in which the magnitude of the immediate error is proportional to the magnitude of the drift and the recovery rate is independent of the magnitude of drift}\label{fig:cvres3}
\end{figure}

Of the tree based classifiers, only \verb?HoeffdingAdaptiveTree? behaves as expected, as shown in Fig.~\ref{fig:cvres3}.

\begin{figure}
\def\addfigcv#1{\hspace*{-6pt}\includegraphics[width=.5\textwidth]{CV#1.pdf}}
\begin{tabular}{cc}
\addfigcv{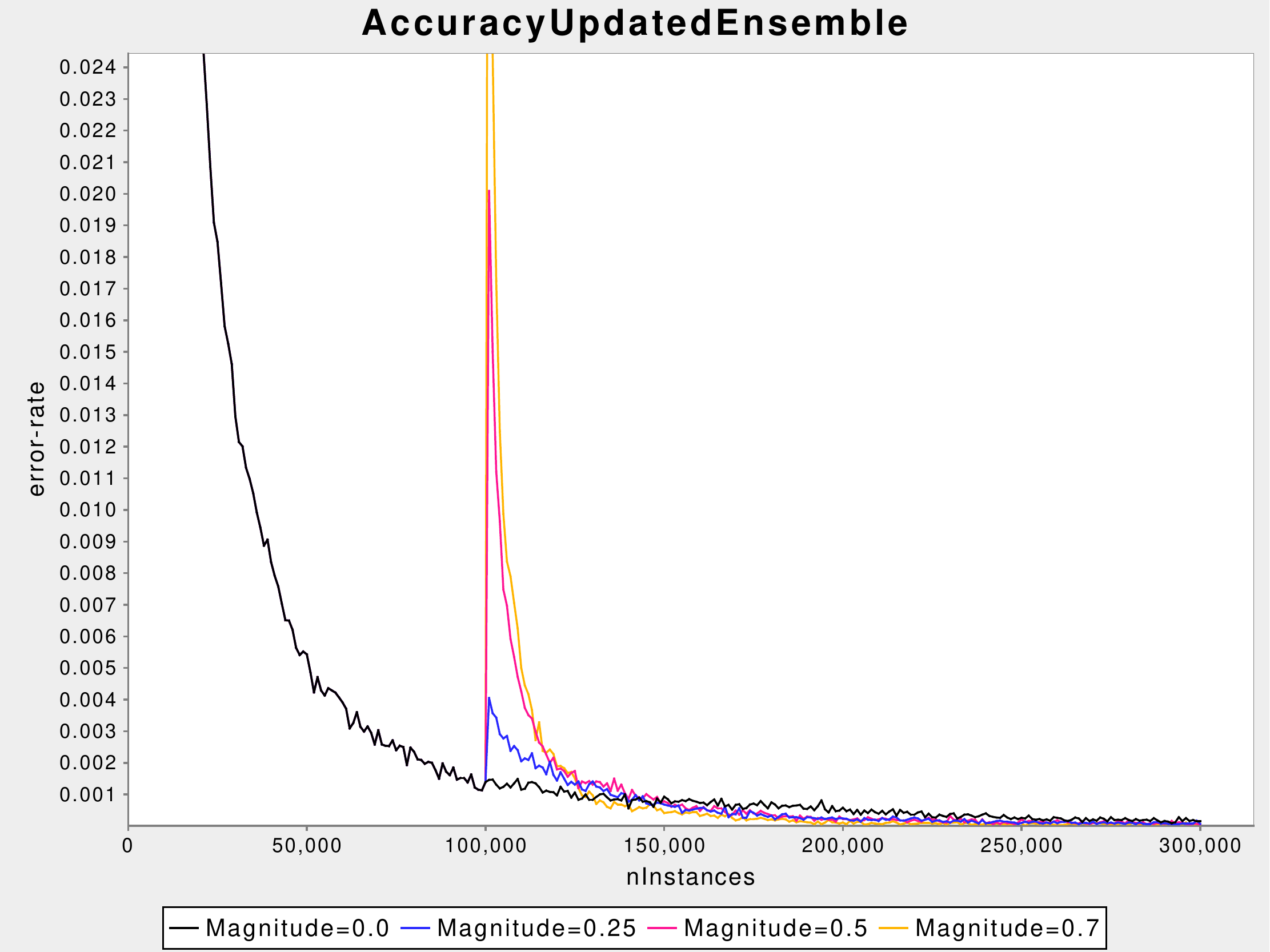}&
\addfigcv{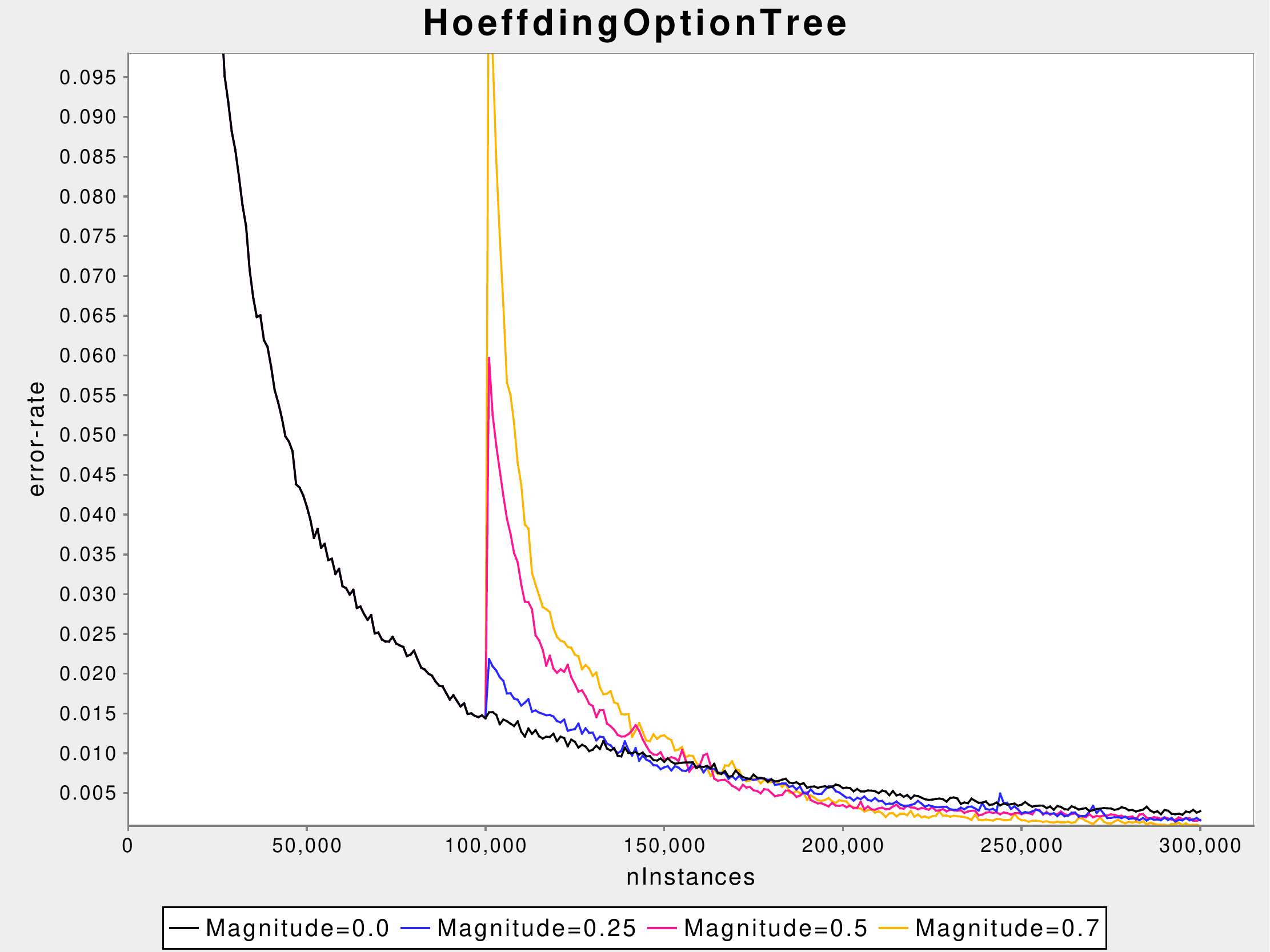}\\
\addfigcv{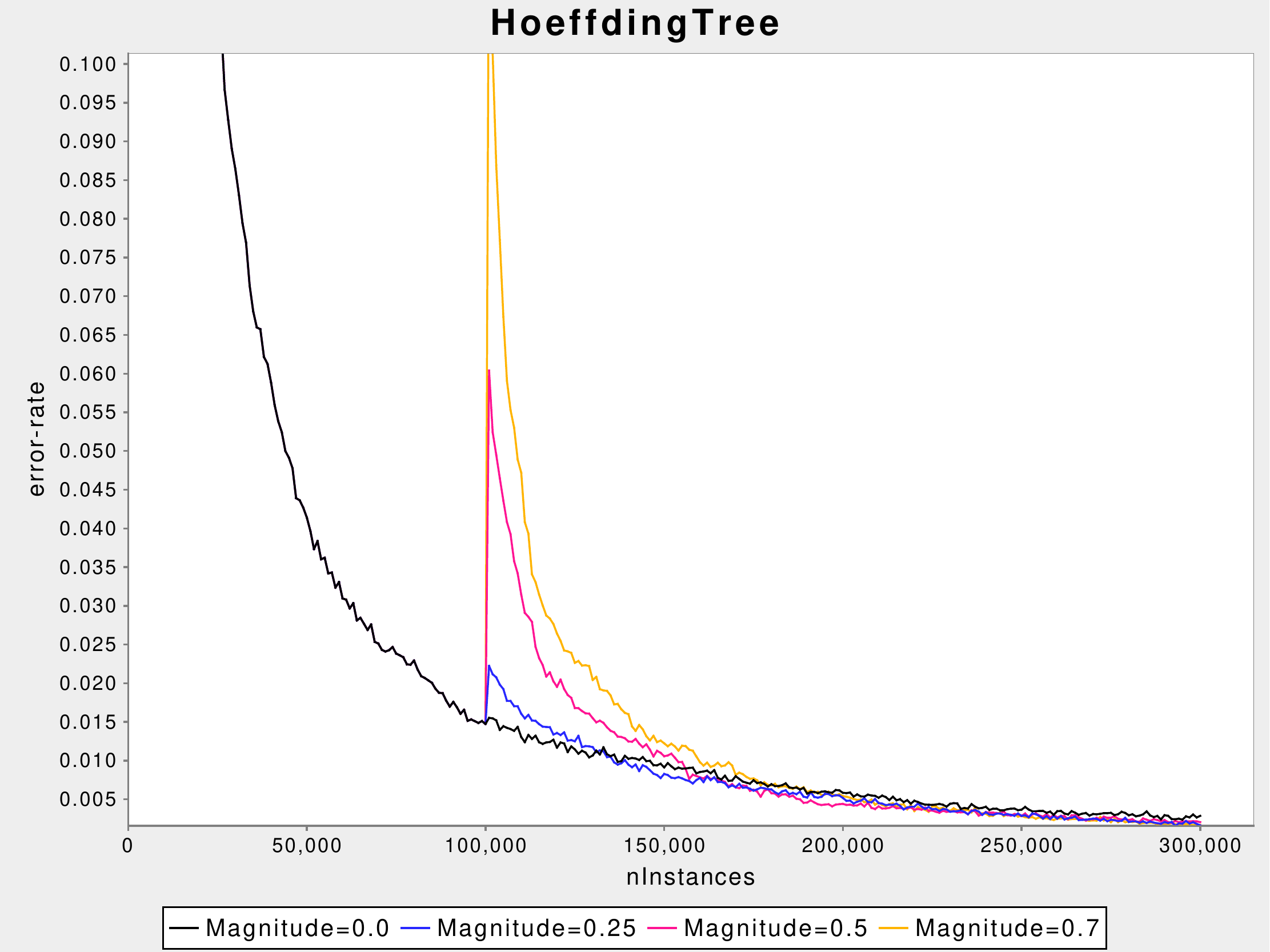}&
\addfigcv{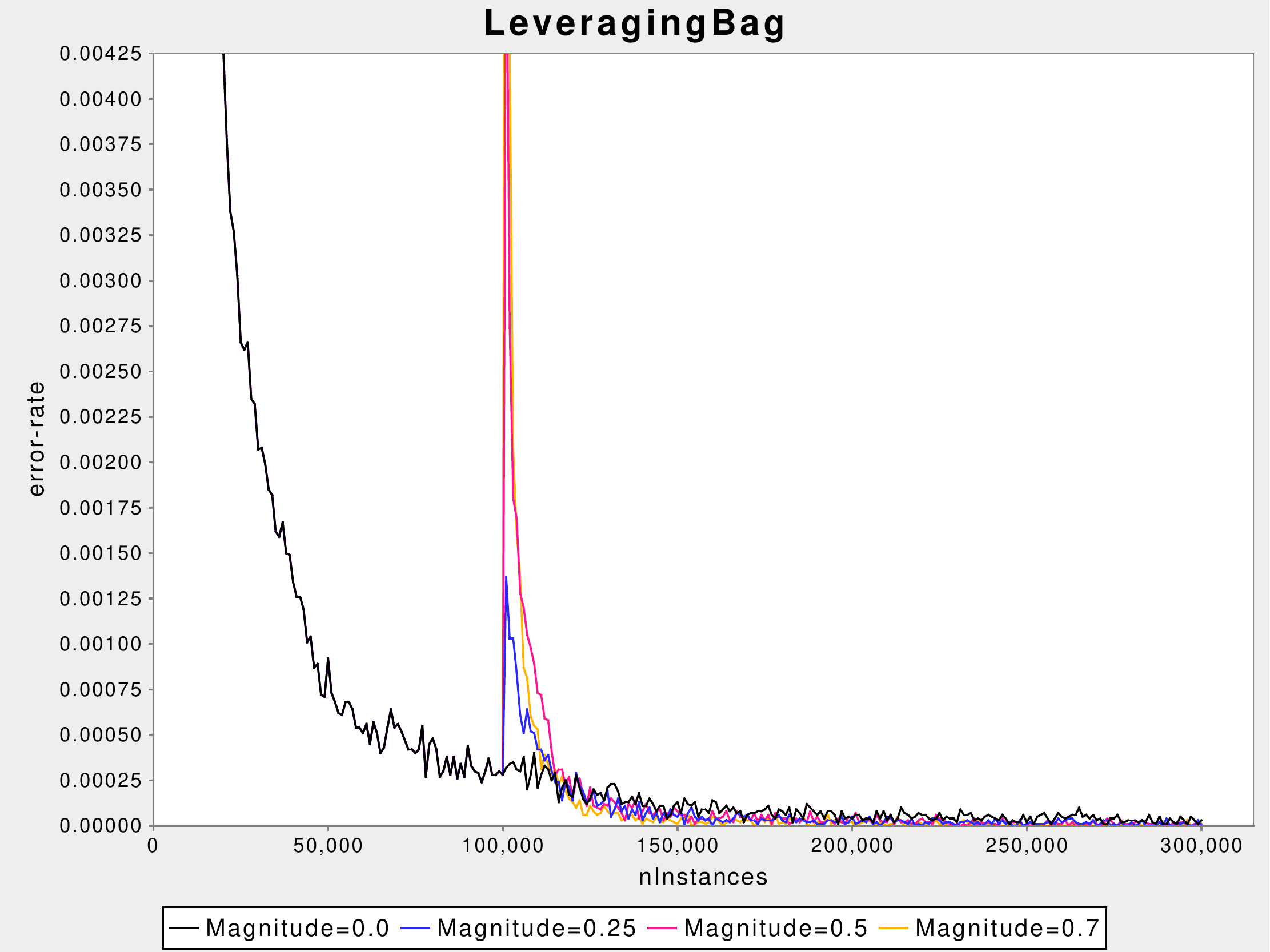}\\
\addfigcv{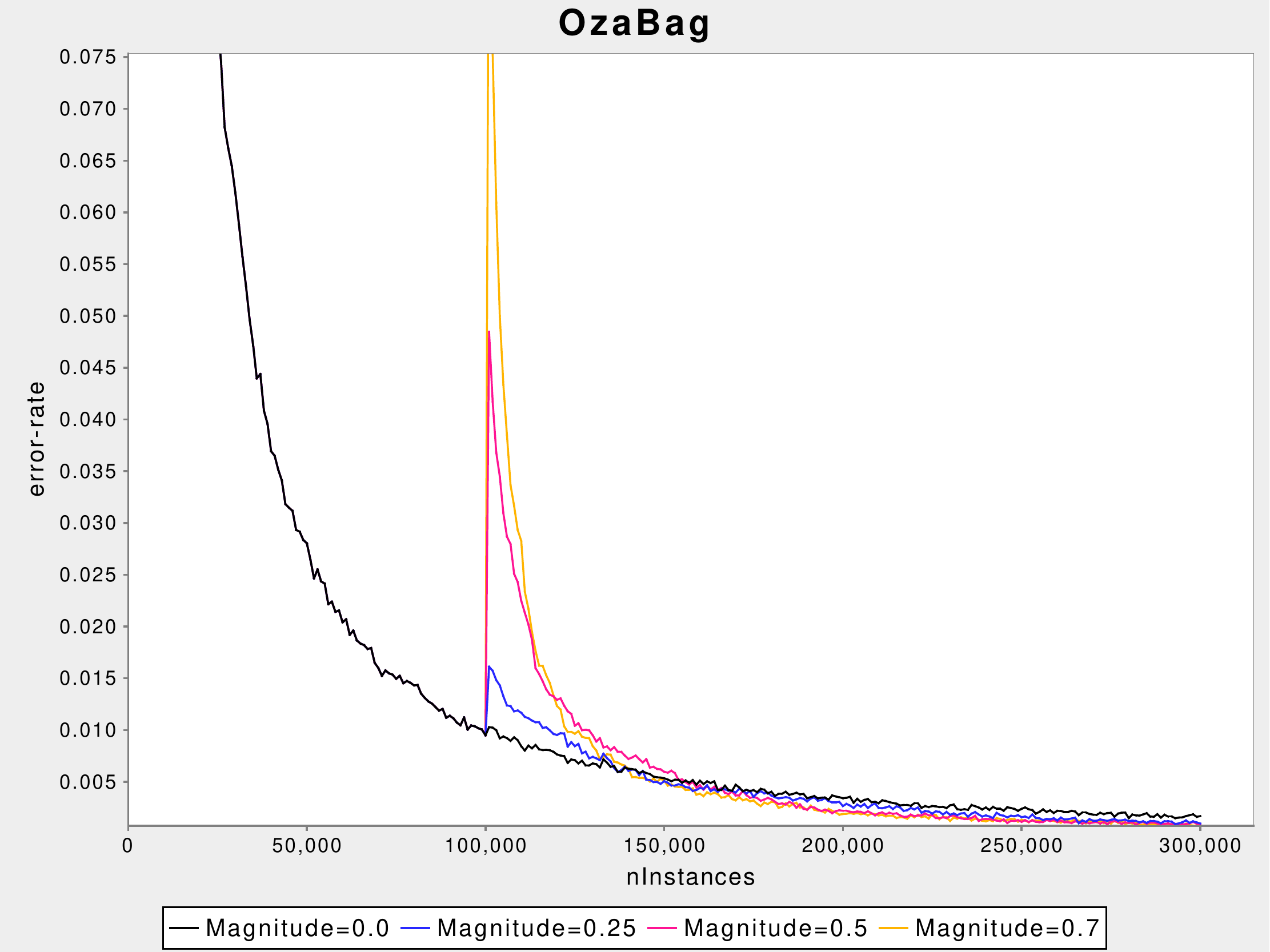}&
\addfigcv{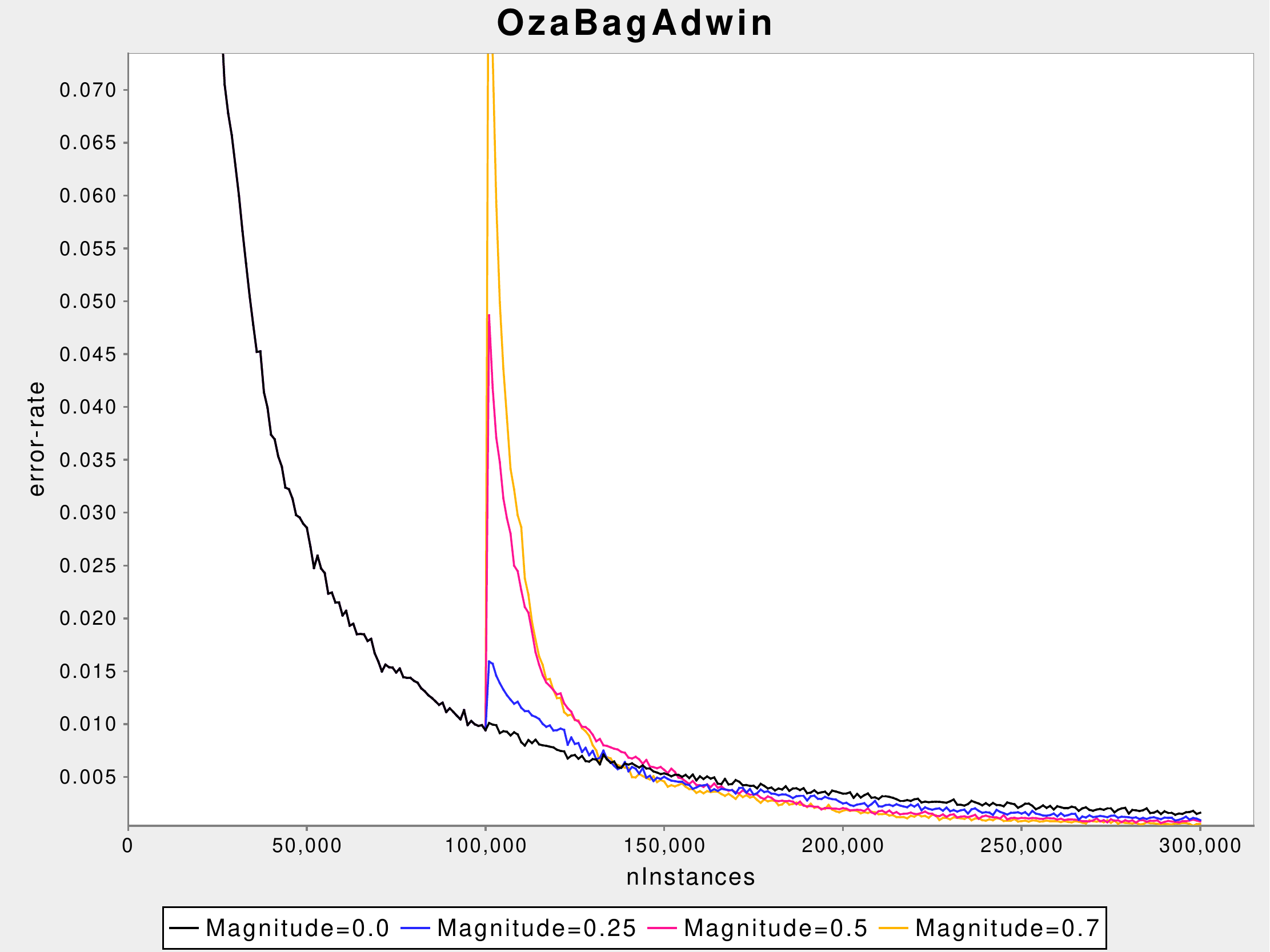}\\
\addfigcv{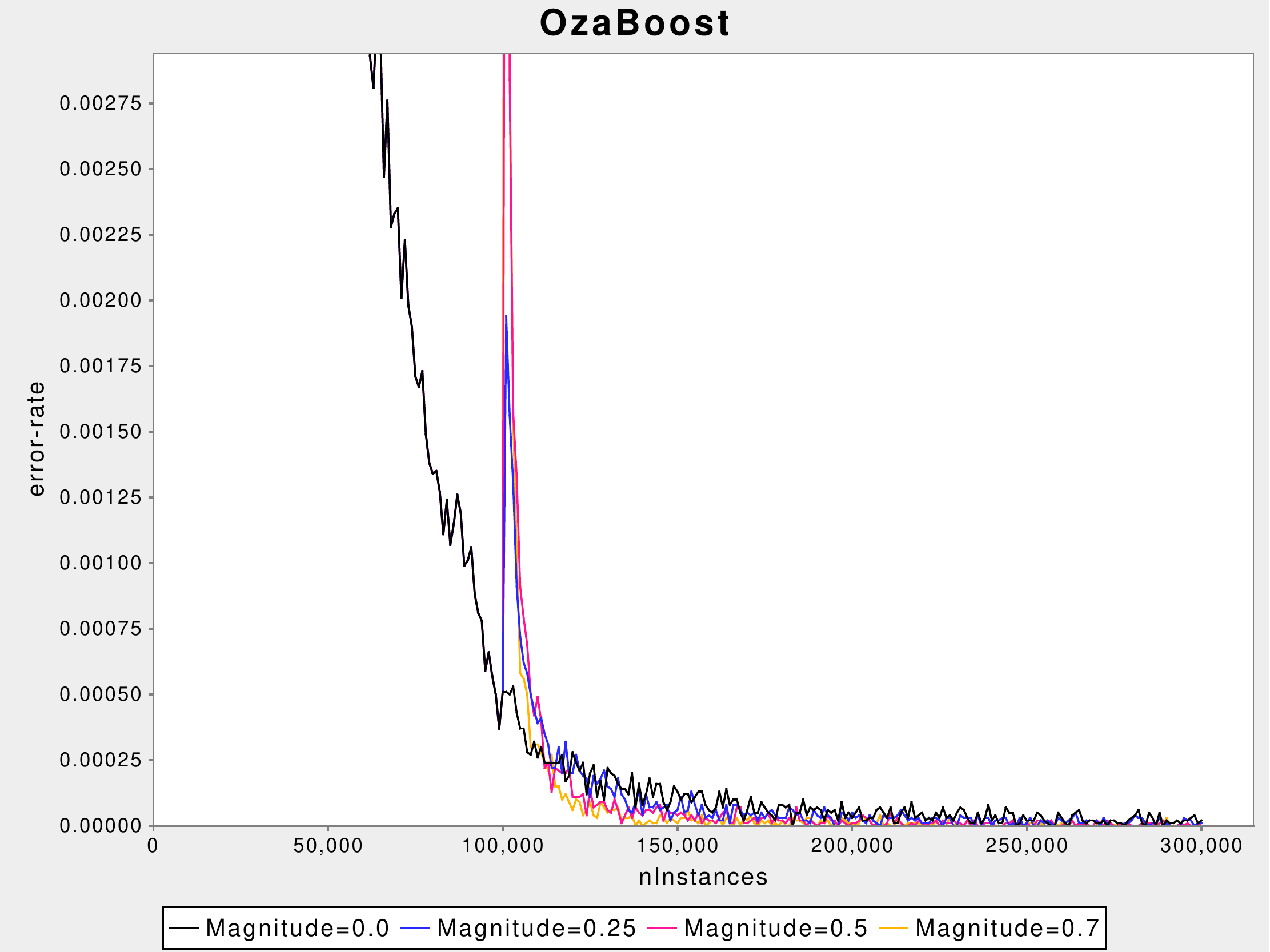}&
\addfigcv{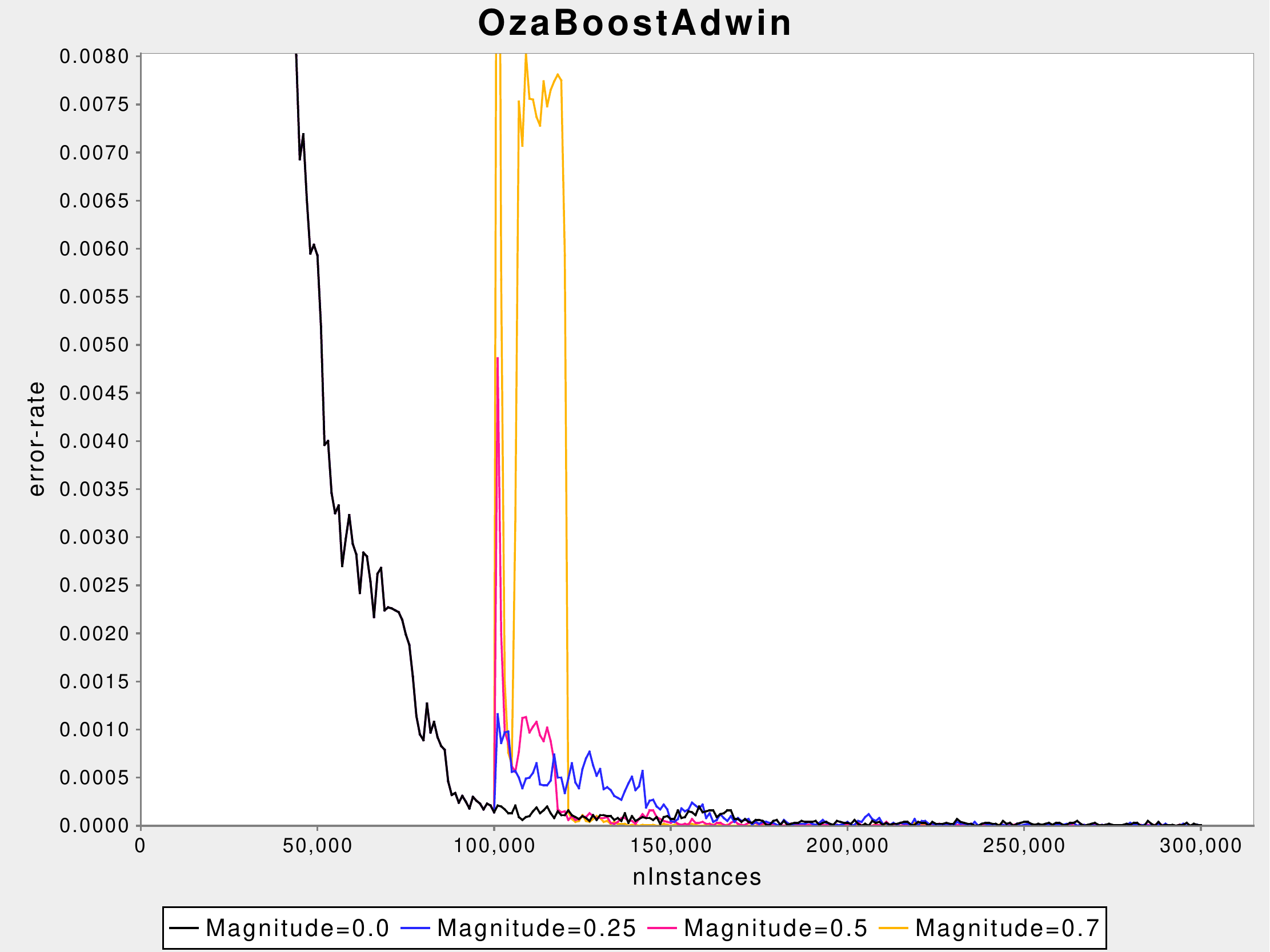}
\end{tabular}
\caption{Pure covariate drift: Learning curves in which larger magnitude covariate drift sometimes results in faster recovery}\label{fig:cvres2}
\end{figure}

The final class of response is shared by the remaining algorithms, for which the greater the magnitude of the drift, the greater the immediate jump in error. A naive expectation is that a tree based classifier should be little affected by pure covariate drift as the form of the ideal model does not change.  However, what is observed is that the covariate drift does produce a substantial jump in error that is proportional to the magnitude of the drift.  Further, the greater the magnitude of the drift, the better the recovery.  Moreover, and even more surprisingly, the larger the magnitude of the drift, the lower the final level of error that is achieved.  The learning curves for these algorithms are shown in Fig.~\ref{fig:cvres2}.

These results are initially quite startling, as it might be expected that a tree based classifier after seeing 100,000 examples would have learned a tree with a leaf for every one of the $3^5=243$ combination of covariate values.  Indeed, in practice slightly fewer than this number of leaves should be required, as, 
by chance, all the leaves below a small number of potential branches should be mapped to the same class and hence the branch should not be included in the tree.

However, the tree learners are clearly not learning a complete tree, as if they had, error would be zero.  This is because some of the combinations of covariate values are, by chance, substantially less frequent than others, and the tree learners fail to detect the value of splitting on them.

The larger the magnitude of a pure covariate drift, the greater the probability that a combination of covariate values that was infrequent prior to the drift will become substantially more frequent after the drift.  The more frequent a combination becomes, the faster the learner will recognize the value of a split to distinguish it.

\begin{figure}
\def\addfigcv#1{\hspace*{-6pt}\includegraphics[width=\textwidth]{CV#1.pdf}}
\subfigure[\label{subfig:zoomed-error}]{\addfigcv{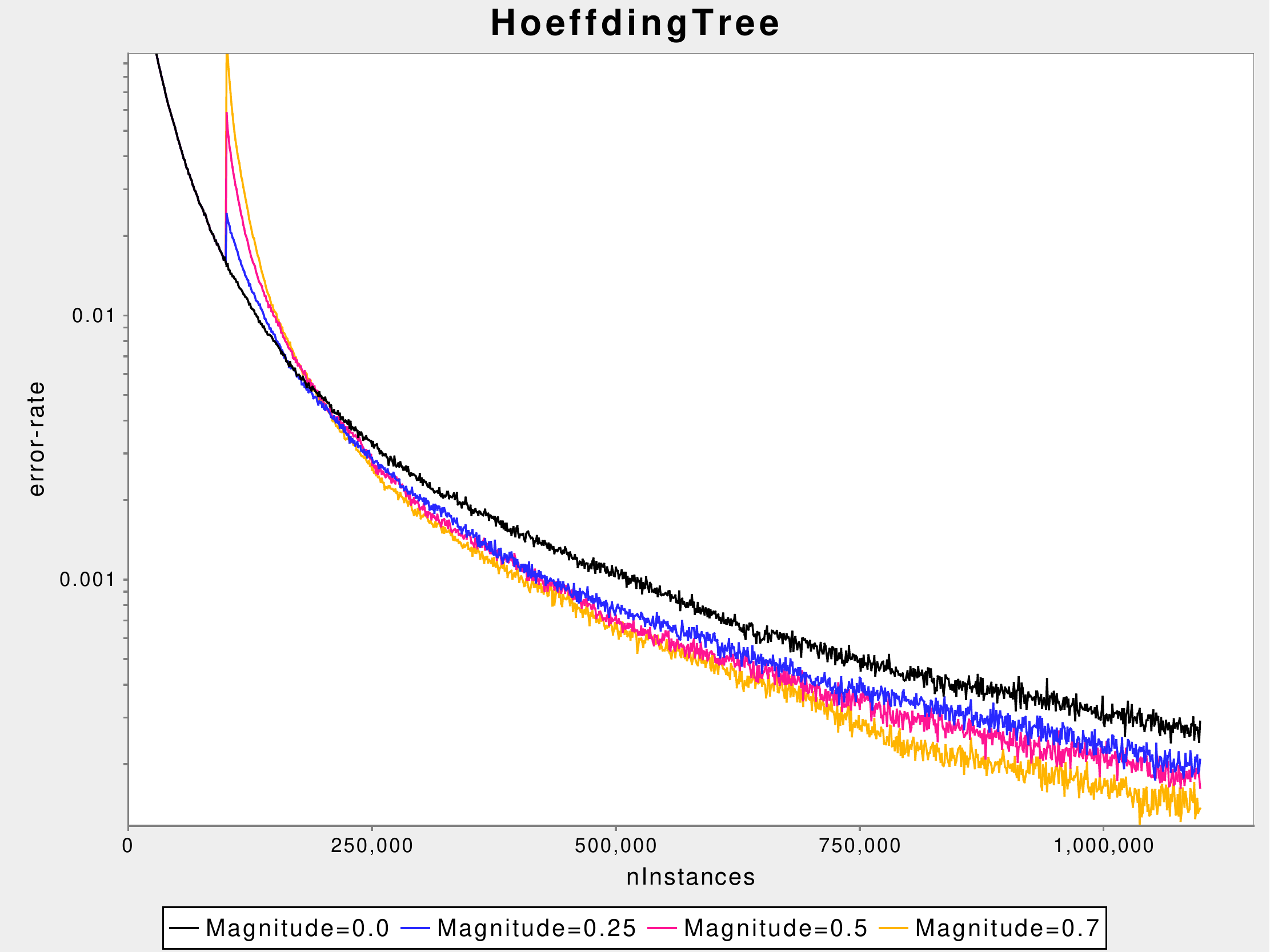}}
\subfigure[\label{subfig:zoomed-WDL}]{\addfigcv{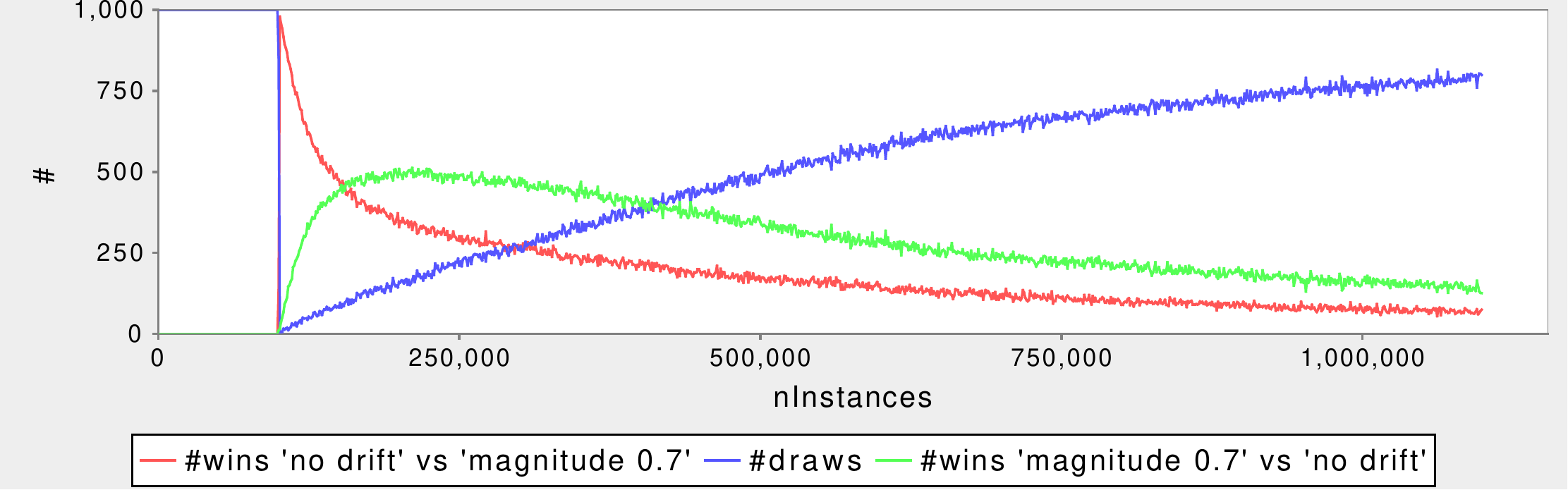}}
\caption{Pure covariate drift: Learning curve for Hoeffding Tree extended to $t=1,100,000$. Y-axis is displayed in log-scale. \subref{subfig:zoomed-error} Error for different magnitude of drift. \subref{subfig:zoomed-WDL} Win-Draw-Loss comparing data streams with no drift to data streams with drift of magnitude 0.7}\label{fig:zoomed}
\end{figure}

To further investigate this phenomenon, we repeated the experiment for the most classic of the tree based learners, the Hoeffding Tree, but continued learning until $t=1,100,000$ (continue learning up to 10 times the number of instances presented before the drift).  We show the resulting curve in Fig.~\ref{subfig:zoomed-error}, using log scale to make clear the consistent but small differences at $t=1,100,000$. Note that the apparent high-variance at the end is only due to the log-scale.
In order to get sufficient resolution to clearly show the differences in error, the experiment was repeated for 1,000 different randomly generated streams.  The results show that even 1,000,000 examples after the pure covariate drift, the larger the magnitude of the drift, the lower the eventual error of the Hoeffding Tree.

The plots of mean error do not reveal how consistent is this advantage to larger magnitudes of drift.  To provide an indication of this, we show in Fig.~\ref{subfig:zoomed-WDL} the Win-Draw-Loss for the setting with no drift (Magnitude=0.0) against the setting with a drift of Magnitude=0.7. The wins for no drift represent the number of data stream pairs for which the error for no drift (averaged over 1,000 successive time steps for the sake of precision) is lower than the error for Magnitude=0.7.  Each data stream-pair includes one data stream generated from the randomly created initial concept (the Magnitude=0.0 stream) and a second stream that starts with the first 100,000 instances from the first stream but then is generated from a randomly formed concept at a Hellinger distance of 0.7 from the initial concept (the Magnitude=0.7 stream). The wins for Magnitude=0.7 represent the number of data stream pairs for which its error is lower, and \#draws indicates the number of stream pairs for which the errors are identical.

Fig.~\ref{subfig:zoomed-WDL} shows that shortly after the drift, in fact from $t=156,000$, the error for Magnitude=0.7 is lower more often than is the error for no drift. Even 1,000,000 time steps after the drift, where the methods seem to be the closest to each other, the error of no drift is lower for only 75 streams, while Magnitude=0.7 has lower error for 126 streams. We conducted a one-tailed binomial sign test of significance (null hypothesis: when there is not a draw, no drift is as likely or more likely to achieve lower error; alternative hypothesis: when there is not a draw, Magnitude=0.7 is more likely to achieve lower error) with the result $p=0.0001$, which is highly significant at the 0.01 level.

\section{Discussion}

These experimental results should only be regarded as preliminary.  They only show the response of each learner in the context of one specific type of pre and post abrupt drift concept.  Apart from the startling result of covariate drift improving the accuracy of tree based learners, no attempt has been made to understand why the different mechanisms respond as they do.  There is clearly much work to be done to develop a comprehensive understanding of how different drift detection and remediation mechanisms perform in the context of different forms of drift.

However, these simple experiments with different magnitudes and types of abrupt drift and their results provide compelling evidence of the importance of quantitative characterizations of drift for understanding the relative response of different drift detection and remediation mechanisms to different forms of drift, for designing effective drift detection and remediation mechanisms, and ultimately for developing a solid theoretical understanding of learning in the context of concept drift.  Without the notion of drift magnitude and the theoretical tools to measure and manipulate it, it would not be possible to derive the insight that the greater the magnitude of pure covariate drift the faster a Hoeffding Tree recovers and the lower the final error attained.

Having established the importance of quantitative characterizations of drift, we now explore some of the implications of our new conceptual analysis.

Many of the qualitative definitions rely on user defined thresholds {$\phi$}, {$\delta$}, {$\beta$}, {$\gamma$}, $\nu$ and $\mu$.  As a result,  these definitions are inherently subjective.  While this may seem a limitation of our formal definitions for existing qualitative characterizations of concept drift, we believe it is actually a useful contribution to make explicit which aspects of these characterizations are subjective.  

Our analysis has made explicit  the subjective nature of many of the core concept drift types that have been previously proposed.  Rather than relying on these subjective qualitative descriptions, we suggest that it is better to use the new objective quantitative measures that we have defined.  These include cycle duration and drift rate, magnitude, frequency, duration and path length.

It is not clear to what extent it might be feasible to use our definitions to directly quantify real-world cases of concept drift, as it will often not be feasible to accurately estimate the relevant probability distributions at each point in time.  Nonetheless, they are likely to be useful in helping the design of drift detection systems \citep{gama2004learning,dries2009adaptive,nishida2007detecting,wang2013concept}, as they provide clear guidance about the ways in which drift may occur. Further, the measures that we define may provide useful idealizations to guide the development of approximations that are feasible to assess from data.  This is another promising direction for follow-up research.

Our definitions should play an important role in informing the design of online learning algorithms.  
By improving understanding of the drift mechanisms that a learner may confront, our definitions have the potential to assist the design of more robust and effective online learning systems.  

Additionally, by providing a comprehensive taxonomy of types of drift and measures for exactly quantifying them, our conceptual analysis should assist the design of more thorough experimental evaluations of the relative capacities of different learners to respond to different forms of drift. For example, synthetic data streams are frequently used to assess relative performance of different algorithms under various possible drift types \citep{Zliobaite2014Perm,brzezinski2014prequential,Shaker2015recovery}. Our taxonomy and definitions of drift types provide comprehensive and structured guidance for the design of synthetic data stream experiments.  They make explicit the different types of drift that may affect algorithm performance.  

Further, they should play an important role in improving the analysis of experimental performance.  They provide a comprehensive set of formal definitions and the first set of quantitative measures of different types of drift.  As a result, they supply a framework for more objective evaluation of stream mining algorithms and their relative capacities to respond to different drift scenarios.

While only a few of the types of drift that we have defined are inherently predictable (see Section~\ref{sec:predictability}), there may be other types of factors that make some aspects of drift predictable.  For example, in some applications it may be possible to detect triggers for different types of drift.  Here our definitions will provide a useful framework for characterizing and exploiting the predictable elements of the drift.

Our formal  qualitative definitions of types of drift are based on the model of a stream as comprising periods of concept stability interspersed by periods of drift.  This model appears to be implicit in many previous characterizations of types of concept drift.  This is a questionable assumption for many streams and one which would be an interesting subject for empirical study. 

\section{Conclusion}

This research began as an exercise in developing formal definitions for existing qualitative descriptions of concept drift. This revealed several shortcomings of qualitative descriptions of drift:
\begin{itemize}
\item that they require subjective thresholds --- the parameters $\phi$,
$\delta$, $\beta$, $\gamma$, $\nu$ and $\mu$ in our definitions;
\item that they appear to rely on an implicit assumption that drift occurs between periods without drift; and
\item that they cannot provide fine-grained discrimination between different drifts of the same general types.
\end{itemize}
These factors highlight the need for quantitative measures of drift.  Only quantitative measures can provide objectivity and precision. To this end we have defined the key measures drift magnitude and duration, from which it is possible to derive further measures, such as rate and path length.



We anticipate that defining quantitative measures for characterizing concept drift and resolving the ambiguities and imprecision of past qualitative definitions will help:
\begin{itemize}
\item standardize the terminology used in the literature;
\item provide a rigorous theoretical basis for designing new mechanisms to detect, characterize and resolve concept drift;
\item understand what forms of drift are best handled by each different mechanism for handling drift;
\item create greater diversity in the types of drift that are created by synthetic drift data generators \citep{Narasimhamurthy07} for stream mining evaluation; 
\item develop more comprehensive and objective approaches for evaluating stream mining algorithms; and
\item  design new machine learning techniques that are robust to a greater diversity of drift types.
\end{itemize}

\section*{Acknowledgments}
We are grateful to David Albrecht, Mark Carman, Bart Goethals, Nayyar Zaidi and the anonymous reviewers for valuable comments and suggestions.

This research has been supported by the Australian Research Council under grant DP140100087 and Asian Office of Aerospace Research and Development, Air Force Office of Scientific Research under contract FA2386-15-1-4007. 

\bibliographystyle{plainnat}

\end{document}